\documentclass[10pt,twocolumn,letterpaper]{article}

\usepackage[pagenumbers]{cvpr} 

\usepackage{multirow}
\usepackage{booktabs}
\usepackage{siunitx}
\usepackage{makecell}
\usepackage{graphicx} 
\usepackage[utf8]{inputenc} 
\usepackage[T1]{fontenc}    
\usepackage{url}            
\usepackage{booktabs}       
\usepackage{amsfonts}       
\usepackage{pifont}
\newcommand{\xmark}{\ding{55}}
\newcommand{\cmark}{\ding{51}}
\usepackage{multirow}       
\usepackage{nicefrac}       
\usepackage{microtype}      
\usepackage{xcolor}         
\usepackage{graphicx}
\usepackage{caption}
\usepackage{tikz}          
\usepackage{pgfplots}      
\usepgfplotslibrary{polar} 
\usepackage{xspace}
\usepackage{amsfonts}       
\usepackage{amsmath}        
\usepackage{algpseudocode}
\usepackage{algorithm}
\usepackage{makecell}
\usepackage{amssymb} 
\usepackage{float}
\usepackage{subcaption}
\usepackage{enumitem} 
\usepackage{listings}

\usepackage[most]{tcolorbox}    
\definecolor{myred}{RGB}{255,153,153}
\definecolor{darkred}{RGB}{165,42,42}

\usepackage{booktabs}
\usepackage[table]{xcolor}

\usepackage{booktabs}
\usepackage{subcaption}
\usepackage{caption} 
\usepackage{adjustbox}

\usepackage{colortbl}
\usepackage{xcolor}
\definecolor{best}{RGB}{205,255,205}    
\definecolor{second}{RGB}{255,255,205}  

\def\eg{\emph{e.g}\onedot} 
\def\ie{\emph{i.e}\onedot}

\definecolor{mygray}{gray}{.9}
\usepackage{subcaption}

\definecolor{cvprblue}{rgb}{0.21,0.49,0.74}
\usepackage[pagebackref,breaklinks,colorlinks,allcolors=cvprblue]{hyperref}

\title{Flowing from Reasoning to Motion: Learning 3D Hand Trajectory Prediction from Egocentric Human Interaction Videos
}

\author{
Mingfei Chen\textsuperscript{1,2} \:
Yifan Wang\textsuperscript{1} \:
Zhengqin Li\textsuperscript{1} \:
Homanga Bharadhwaj\textsuperscript{1} \:
Yujin Chen\textsuperscript{1} \:
Chuan Qin\textsuperscript{1} \\
Ziyi Kou\textsuperscript{1} \,
Yuan Tian\textsuperscript{1} \,
Eric Whitmire\textsuperscript{1} \,
Rajinder Sodhi\textsuperscript{1} \,
Hrvoje Benko\textsuperscript{1} \,
Eli Shlizerman\textsuperscript{2} \,
Yue Liu\textsuperscript{1} \\ \\
\textsuperscript{1}\textbf{Meta}\qquad
\textsuperscript{2}\textbf{University of Washington} \\ \\
\url{https://egoman-project.github.io/}\\
}

\begin{document}

\twocolumn[{
\renewcommand\twocolumn[1][]{#1}
\maketitle
\begin{center}
\vspace{-25pt}
    \includegraphics[width=2.0\columnwidth,keepaspectratio]{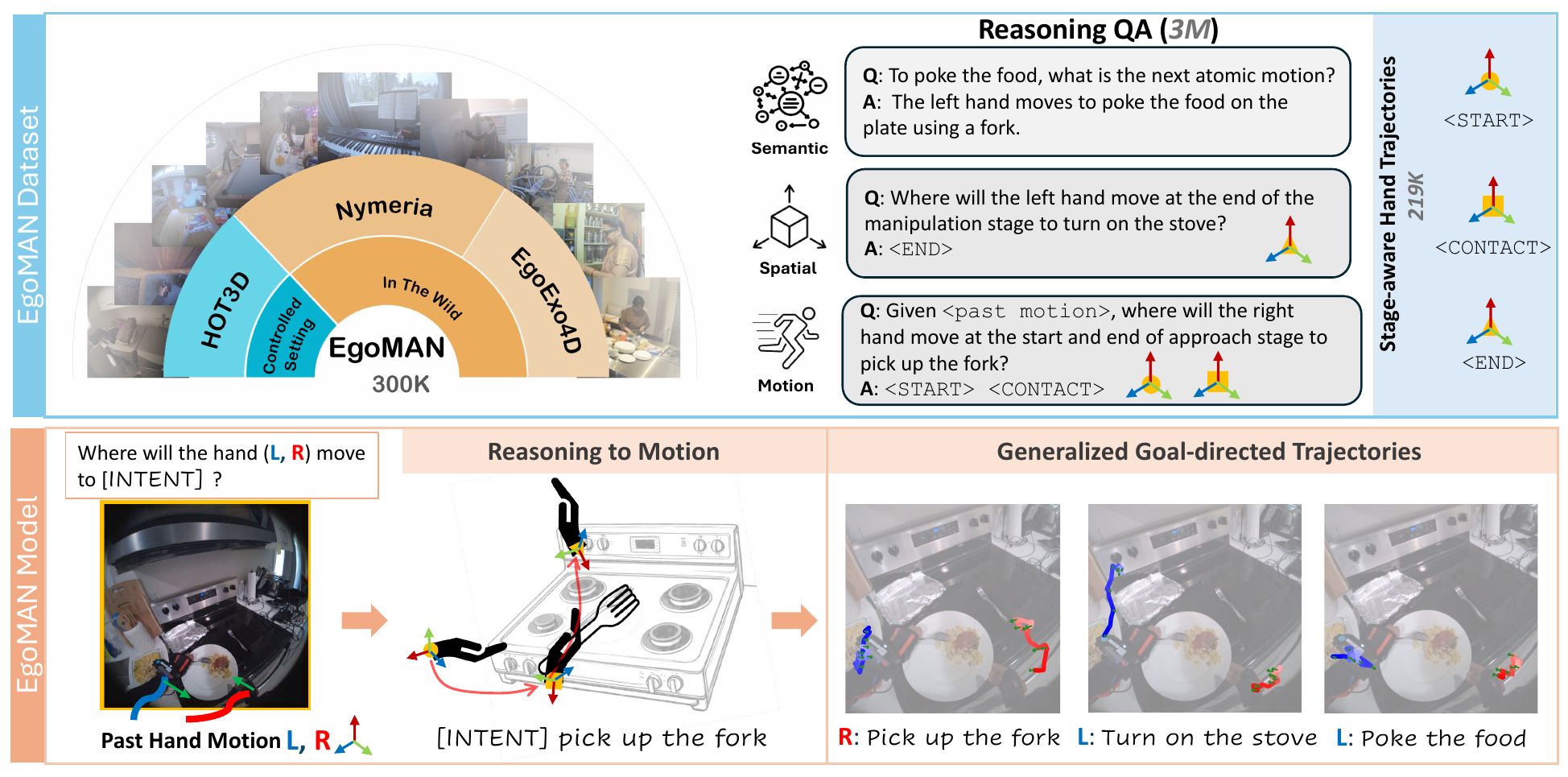}
    \vspace{-7pt}
    \captionof{figure}{
    \textbf{EgoMAN project.} We introduce 1) the \textbf{EgoMAN dataset} (top), a large-scale egocentric dataset for interaction stage–aware 3D hand trajectory prediction with 219K 6-DoF trajectories and 3M structured QA pairs for semantic, spatial, and motion reasoning. During inference, 2) the \textbf{EgoMAN model} (bottom) takes an image, past hand motion, and an intent query as input, performs stage-aware reasoning to infer intent-specific waypoints, and then generates 6-DoF hand trajectories of distinct motions for different intent queries.
    }
    \vspace{-5pt}
    \label{fig:teaser}
\end{center}
}]

\maketitle

\begin{abstract}
Prior works on 3D hand trajectory prediction are constrained by datasets that decouple motion from semantic supervision and by models that weakly link reasoning and action.
To address these, we first present the \textbf{EgoMAN dataset}, a large-scale egocentric dataset for interaction stage–aware 3D hand trajectory prediction with 219K 6DoF trajectories and 3M structured QA pairs for semantic, spatial, and motion reasoning. 
We then introduce the \textbf{EgoMAN model}, a reasoning-to-motion framework that links vision–language reasoning and motion generation via a trajectory-token interface. 
Trained progressively to align reasoning with motion dynamics, our approach yields accurate and stage-aware trajectories with generalization across real-world scenes.
\end{abstract}
    
\section{Introduction}
\label{sec:intro}

\noindent Predicting future 3D hand motion is essential for \emph{in-context interaction} and proactive assistance, where a system anticipates human intent from visual, linguistic, and motion cues. Humans perform this naturally, \ie, understanding the goal of an action, interpreting the scene layout, and coordinating movement based on recent dynamics. Achieving this computationally requires jointly reasoning about task semantics, spatial geometry, and temporal motion. We develop a model that predicts long-horizon 3D hand trajectories by integrating visual and motion context with language, which conveys intent and disambiguates visually similar actions. Such capabilities enable applications in robot manipulation, language-conditioned motion synthesis, and assistive systems that respond to human intent.

A major bottleneck is the lack of large-scale, high-quality 3D trajectory data. Controlled datasets~\cite{hot3d, H2O, egopat3d} offer accurate annotations but limited diversity, while large-scale egocentric video datasets~\cite{egoexo4d, nymeria} contain rich real-world interactions but noisy, weakly goal-directed trajectories and little temporal structure. Crucially, they lack explicit \textbf{interaction stages}, \eg, \emph{approach} and \emph{manipulation}, which are needed to separate purposeful motion from background and to connect trajectories to intent. Models trained on such raw videos often generalize poorly because the links between intent, spatial relations, and motion dynamics are missing.

Beyond data limitations, existing modeling approaches also fall short. Affordance-based methods~\cite{vrb, chen2025vidbot} rely on object detectors and affordance estimators, which propagate upstream detection errors and introduce additional computational overhead. End-to-end motion predictors, including those based on diffusion~\cite{diffip2d, Hatano2025EgoH4}, variational~\cite{bao2025handsonvlm}, and state-space models~\cite{BaoUSST_ICCV23}, focus on short-term dynamics with limited semantic grounding. Vision-Language-Action (VLA) systems~\cite{kim24openvla,zitkovich2023rt,luo2025being} exhibit strong reasoning ability, but applying VLMs~\cite{qwen2025qwen25technicalreport, palm-e, chen2024internvl, liu2023visualinstructiontuningllava, grattafiori2024llama3herdmodels} directly to generate continuous 3D motion remains challenging, as they struggle to produce smooth, high-frequency action sequences. Bridges between VLM reasoning and motion experts~\cite{lipman2023flow, pertsch2025fast, black2024pi0, physicalintelligence2025pi05, wen2025dexvla, lee2025molmoact, li2025hamster, embodied-reasoner} typically rely on implicit tokens or lengthy reasoning chains, which limits efficiency, generalization, and interpretability when generating fine-grained, fast actions.

To address these challenges, we introduce the \textbf{EgoMAN} project, which couples a large-scale, stage-aware dataset with a modular reasoning-to-motion framework. The \textbf{EgoMAN dataset} contains over 300K egocentric clips from 1,500+ scenes, including 219K 6DoF hand trajectories annotated with interaction stages (\emph{approach}, \emph{manipulation}) and 3M structured vision–language–motion QA pairs. This supervision explicitly encodes \emph{why}, \emph{when}, and \emph{how} hands move, enabling models to learn intent-linked, spatially grounded motion patterns at scale.

Building on this dataset, the \textbf{EgoMAN model} introduces a compact \textbf{trajectory-token interface} that connects high-level reasoning to continuous 3D hand motion. We define four trajectory tokens: one semantic token (\texttt{<ACT>}) and three stage-aware waypoint tokens (\texttt{<START>}, \texttt{<CONTACT>}, \texttt{<END>}) marking key transitions in interaction. These tokens represent wrist-centered spatio-temporal waypoints rather than object-centric affordances, providing a clear, structured interface for conditioning a flow-matching motion expert. A progressive three-stage training strategy learns (i) intent-conditioned and stage-aware reasoning over semantics, spatial and motion, (ii) motion dynamics, and (iii) their alignment through the token interface, enabling long-horizon, intent-consistent 3D trajectory prediction in diverse real-world scenes.

\vspace{1mm}
\noindent Our main contributions are:
\begin{itemize}
\item \textbf{EgoMAN dataset}: a large-scale, interaction stage–aware 6DoF hand trajectory dataset with structured semantic, spatial, and motion reasoning annotations.
\item \textbf{EgoMAN model}: a modular reasoning-to-motion architecture with a trajectory-token interface and progressive training that aligns semantic intent with physically grounded motion generation.
\item We achieve state-of-the-art accuracy and generalization with high efficiency in 3D hand trajectory prediction across diverse real-world egocentric scenes.
\end{itemize}

\section{Related Works}
\label{related}
\noindent\textbf{Hand Trajectory Prediction.}
Egocentric hand forecasting aims to infer future hand motion from past observations under ego-motion and depth ambiguity. Large-scale works often predict short-horizon \emph{2D} trajectories at low framerates~\cite{liu2022joint, Hatano2024EMAG, diffip2d, ma2024madiff, bao2025handsonvlm}, while curated datasets enable \emph{3D} trajectory prediction~\cite{BaoUSST_ICCV23, ma2025mmtwin, Hatano2025EgoH4}.  
Prior 3D methods generally follow either:  
(a) \emph{object-centric, affordance-driven models}~\cite{liu2022joint, vrb, chen2025vidbot}, which rely on detectors and affordance estimators but suffer from error propagation and additional computational efficiency cost from detection;  
or (b) \emph{end-to-end motion models} predicting trajectories directly from video and past hand motion~\cite{ma2025mmtwin, ma2024madiff, diffip2d}, sometimes incorporating egomotion~\cite{ma2025mmtwin, ma2024madiff, diffip2d} or 3D priors such as point clouds~\cite{ma2025mmtwin}.  
Given that 3D labels are often uncertain and scarce~\cite{BaoUSST_ICCV23}, generative models have become standard: VAEs~\cite{bao2025handsonvlm}, state-space models~\cite{BaoUSST_ICCV23}, diffusion~\cite{diffip2d, Hatano2025EgoH4}, and hybrid variants~\cite{ma2024madiff, ma2025mmtwin}.  
However, these methods typically forecast short fixed horizons, focus on low-level motion, and encode intent implicitly, limiting generalization in diverse real-world egocentric scenarios. 
Our work instead predicts long-horizon, semantically grounded 6DoF trajectories by explicitly conditioning on intent, spatial context, and interaction stages.

\noindent\textbf{Learning Interactions from Human Videos.}
Human videos provide rich demonstrations of hand–object interactions, driving research in reconstruction and forecasting~\cite{liu2020fhoi, liu2022joint, bharadhwaj2024towards, BaoUSST_ICCV23, diffip2d}.  
Controlled datasets~\cite{hot3d, H2O, egopat3d} offer precise 3D annotations but limited task diversity; robotic imitation datasets~\cite{HoloAssist2023, kareer2025egomimic, hoque2025egodex} provide structured demonstrations but remain narrow and scripted.  
Large-scale egocentric datasets~\cite{egoexo4d, nymeria} capture varied daily activities with language annotations but often contain noisy trajectories and unclear interaction boundaries.  
We address these gaps by curating \emph{EgoMAN-Bench}, consolidating real-world egocentric datasets into a stage-aware supervision benchmark. Our model builds on this benchmark to connect reasoning about interaction stages with accurate, long-horizon 3D trajectory prediction, aligning with recent efforts in robot learning from human videos~\cite{bharadhwaj2024gen2act, vrb, track2act, generalflow, kareer2025egomimic, hoque2025egodex}.


\noindent\textbf{Vision-Language Models for Embodied AI.}
Modern VLMs unify perception and language~\cite{liu2023visualinstructiontuningllava, maaz2023videochatgpt, hanoona2023GLaMM, lai2024lisareasoningsegmentationlarge, palm-e, qwen2025qwen25technicalreport, gemmateam2025gemma3technicalreport}, enabling broad video understanding and reasoning.  
Their extensions to Vision-Language-Action (VLA) systems~\cite{kim24openvla, zitkovich2023rt, luo2025being} support manipulation and navigation via robot datasets, but direct action prediction through VLMs often struggles to produce smooth, high-frequency trajectories.  To mitigate this, recent works have sought to incorporate hand trajectory prediction in VLAs either through pre-training or co-training~\cite{yang2025egovla,vitra}. Coupled VLM–motion systems, where the VLM is linked to an action module, use implicit feature routing~\cite{black2024pi0, physicalintelligence2025pi05, wen2025dexvla}, which suffers from poor generalization and limited interpretability, while other approaches rely on long reasoning chains as the interface~\cite{lee2025molmoact, li2025hamster, embodied-reasoner}, resulting in high inference cost and low efficiency.
In contrast, we introduce a trajectory-token interface that directly links high-level reasoning to continuous 3D motion using four specialized semantic and spatiotemporal waypoint tokens, enabling an efficient, interpretable interface that effectively guides the motion expert to generate smooth, accurate high-frequency trajectories.

\section{EgoMAN Dataset}
\label{sec:benchmark}

\noindent
EgoMAN is a large-scale egocentric interaction dataset (300+ hrs, 1{,}500+ scenes, 220K+ 6DoF trajectories) built from Aria glasses~\cite{engel2023project} across EgoExo4D~\cite{egoexo4d}, Nymeria~\cite{nymeria}, and HOT3D-Aria~\cite{hot3d}. It provides high-quality wrist-centric trajectories, structured interaction annotations, and rich QA supervision for semantic, spatial, and motion reasoning. This section summarizes dataset statistics, annotation pipeline, trajectory annotation, QA construction, and the dataset split.

\noindent\textbf{Dataset Statistics.}
The data spans diverse interactions—scripted manipulation in HOT3D-Aria, real-world activities (bike repair, cooking) in EgoExo4D, and everyday tasks in Nymeria. Trajectories cover substantial variation: 27.8\% exceed 2\,s, 34.0\% move over 20\,cm on average, and 35.3\% rotate more than $60^\circ$.  
We train on EgoExo4D and Nymeria and reserve HOT3D-Aria as test-only set.

\vspace{1mm}
\noindent\textbf{Annotation Pipeline.}
We use GPT-4.1~\cite{openai2024gpt4technicalreport} to extract interaction annotations for EgoExo4D and Nymeria. At each atomic action timestamp~\cite{egoexo4d,nymeria}, we crop a 5\,s clip and annotate two wrist-centric interaction stages:  
(1) \textit{Approach}—the hand moves toward the target manipulation region;  
(2) \textit{Manipulation}—the hand performs the action with the object in hand.
Detailed prompts and filters are provided in the appendix.
For HOT3D, we infer interaction stages using hand–object trajectories, defining approach as 0.5–2.0\,s prior to object motion (object visible and within 1\,m), and the manipulation stage corresponds to period after motion onset.

\noindent\textbf{EgoMAN Trajectory.}  
The EgoMAN dataset provides 6DoF wrist trajectories for both hand wrists (3D position + 6D rotation~\cite{zhou2019continuity}), sampled at 10 frames per second (FPS).  
For \textit{EgoExo4D}, we use hand tracking data produced by Aria’s Machine Perception Services (MPS)~\cite{mps}.    
For Nymeria dataset, we use trajectories obtained from two wrist-mounted devices.  
For HOT3D dataset, we directly use the high-quality 6DoF hand trajectories provided by the dataset.
All trajectories are aligned by transforming positions and orientations into the camera coordinate frame of the final visual frame before interaction begins.
\vspace{1mm}

\noindent\textbf{EgoMAN QA.}
We generate structured question--answer pairs using GPT, covering \textit{semantic} (21.6\%), \textit{spatial} (42.6\%), and \textit{motion} (35.8\%) reasoning .

\noindent(1) \textit{Semantic reasoning} questions target high-level intent, 
such as:
\begin{itemize}
    \item “\textit{What will be the next atomic action?}”
    \item “\textit{What object will the hand interact with next?}”
    \item “\textit{Why does the next action happen?}”
\end{itemize}
These questions connect language to goal-directed hand behaviors, enabling deeper understanding of the motivations and purposes behind specific actions.

\noindent(2) \textit{Spatial reasoning} questions ground intent within metric 3D space by querying the wrist’s state at key interaction stages such as approach onset, manipulation onset (approach completion), and manipulation end. These questions may target a single stage (\eg, “\textit{Where/When will the left hand complete the manipulation?}”) or span multiple stages (\eg, “\textit{Where/When is the right hand at the start and end of manipulation?}”), enabling reasoning about transitions between interaction stages. Some questions explicitly reference objects and stage timestamps, supporting reasoning over object-time-space relationships that align with interaction intent.

\noindent(3) \textit{Motion reasoning}  questions probe how past motion informs both semantic and spatial understanding, supporting reasoning about the evolution of motion over time. To construct these questions, we augment a random subset of semantic and spatial questions by prepending a 0.5-second 6DoF hand trajectory sequence from before the interaction start time.  (\eg, {"\textit{Given the <past motion>, where will the right hand complete the approach stage?}”})
This approach enables analysis of how previous hand movements influence subsequent actions and spatial positions, deepening the connection between motion history and interaction intent.

\noindent\textbf{Dataset Split.}  
To support our progressive training pipeline, we split the EgoMAN dataset into 1{,}014 scenes for pretraining (64\%), 498 for finetuning (31\%), and 78 for testing (5\%).
The pretraining set contains lower-quality trajectory annotations—where the target object may be occluded, image quality is low, or interaction intent is ambiguous, and interacitons are generally sparse. In total, the pretrain set comprises 74K samples, 1M QA pairs. The finetune set, by contrast, provides 17K high-quality trajectory samples.  

\noindent For evaluation, we introduce \textbf{EgoMAN-Bench} as our test set, which consists of two settings:  
\noindent(1) \textbf{EgoMAN-Unseen} includes 2{,}844 trajectory samples from 78 held-out EgoExo4D and Nymeria scenes with high-quality trajectories, used to evaluate generalization to new in-domain but previously unseen scenes.
\noindent(2) \textbf{HOT3D-OOD}  includes 990 trajectory samples from HOT3D (dataset only used in testing), designed to evaluate out-of-distribution (OOD) performance on novel subjects, objects, and environments.

\begin{figure*}[t]
\vspace{-0.2in}
    \centering
    \includegraphics[width=0.9\linewidth]{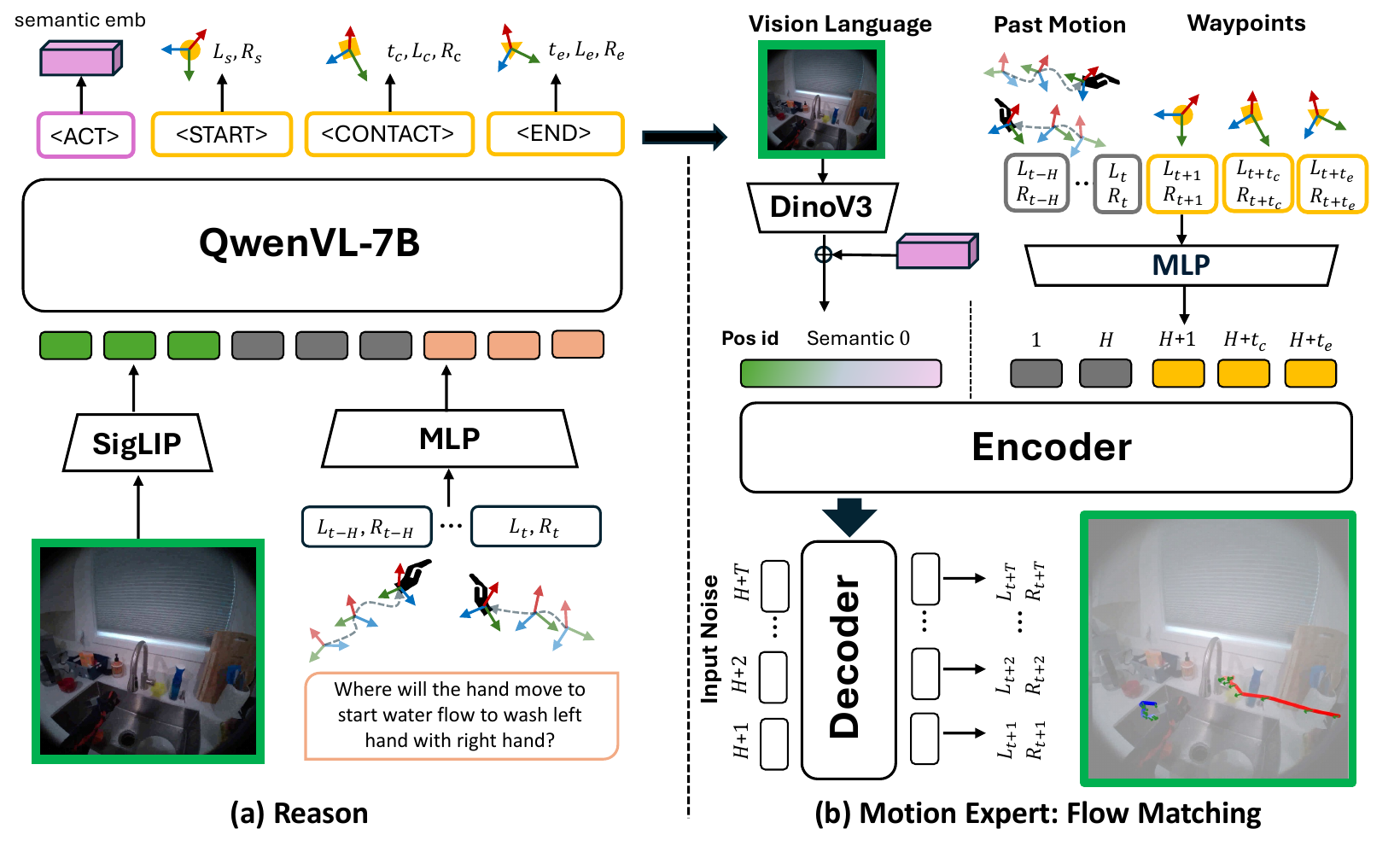}
    \captionsetup{aboveskip=-0.25mm}
\caption{\textbf{Overview of the EgoMAN model.} The EgoMAN model is a modular reasoning-to-motion framework that predicts future 6DoF hand trajectories from an egocentric RGB frame, past wrist trajectories, and a language intent. The \textbf{Reasoning Module} (a), built on QwenVL-7B, extracts semantic and spatial features and outputs trajectory tokens with waypoints and intent semantic cues. The \textbf{Motion Expert} (b), using Flow Matching, predicts future trajectories based on waypoints, past motion, intent semantics and visual input. The trajectory tokens of (a) form the \textbf{Trajectory-Token Interface} which replaces semantic and waypoint condition inputs of (b) to bridge from Reasoning to Motion Expert.}
\label{fig:main}
\end{figure*}
\section{EgoMAN Model}
\label{sec:methods}

As illustrated in Fig.~\ref{fig:main}, the \textbf{EgoMAN} model has two components:
a \textbf{Reasoning Module} that extracts cues and reasons over semantics, spatial relations, and motion to produce stage-aware waypoints, and
a \textbf{Motion Expert} that generates 6DoF hand trajectories.
In this section, we first formalize the prediction task in Sec.~\ref{sec:problem_formulation}, then detail the Reasoning Module in Sec.~\ref{sec:reasoning_pre-training}, the Motion Expert in Sec.~\ref{sec:motion_pre-training}, and Reasoning to Motion via Trajectory-Token Interface in Sec.~\ref{sec:joint_training}.

\subsection{Problem Formulation}
\label{sec:problem_formulation}
Given a single RGB frame $\mathbf{V}_t$, past wrist trajectories $\{\mathbf{L}_\tau, \mathbf{R}_\tau\}_{\tau=t-H}^{t}$, and an intent description $\mathbf{I}$ as input, the task is to predict future 6DoF trajectories $\{\Tilde{\mathbf{L}}_\tau, \Tilde{\mathbf{R}}_\tau\}_{\tau=t+1}^{t+T}$ across manipulation stage, \eg, reaching, manipulating, or releasing. Each position vector $\mathbf{L}_\tau \in \mathbb{R}^6$ and rotation vector $\mathbf{R}_\tau \in \mathbb{R}^{12}$ represent the 3D positions and 6D rotations of both wrists.
Our \textbf{EgoMAN} model acts as the function $\mathcal{F}$ that maps the inputs to future trajectories:
\[
\mathcal{F}: \left( \mathbf{V}_t, \{\mathbf{L}_\tau, \mathbf{R}_\tau\}_{\tau=t-H}^{t}, \mathbf{I} \right) \mapsto \{\Tilde{\mathbf{L}}_\tau, \Tilde{\mathbf{R}}_\tau\}_{\tau=t+1}^{t+T}
\]

\subsection{Reasoning Module}
\label{sec:reasoning_pre-training}
To predict accurate hand trajectory that aligns with human intent and environment context, we need to understand the spatial context of the environments as well as intent semantics. Therefore, the first module of EgoMAN model is reasoning model which aligns spatial perception and motion reasoning with task intent semantics and interaction stages.
Built on Qwen2.5-VL~\cite{qwen2025qwen25technicalreport}, it takes as input an egocentric frame $\mathbf{V}_t$, a language query with intent description~$\mathbf{I}$, and past wrist trajectories $\{\mathbf{L}_\tau, \mathbf{R}_\tau\}_{\tau=t-H}^{t}$. The past-motion sequence is encoded into the same latent space as the VLM's visual and language features, and then fused with them. Depending on the query, the module outputs either (i) a natural-language answer or (ii) a set of structured \emph{trajectory tokens} that represent key interaction semantics and waypoints.

We introduce four trajectory tokens, one action semantic token and three waypoint tokens, to explicitly capture intent semantics and key spatiotemporal transitions across interaction stages.
The action semantic token\texttt{<ACT>} decodes an action semantic embedding corresponding to the interaction phrase (e.g., ``left hand grabs the green cup''). 
The three waypoint tokens: \texttt{<START>}, \texttt{<CONTACT>}, and \texttt{<END>} denote the approach onset, manipulation onset (\ie, approach completion), and maunipulation completion stages respectively. Each waypoint token is equipped with a lightweight head that predicts a timestamp, 3D wrist positions, and 6D wrist rotations. These tokens allow the module to align semantic intent with the corresponding spatiotemporal hand states. 

\paragraph{Reasoning Pre-training.}
To support this dual functionality to predict text and trajectory tokens, we first pretrain the module on 1M question--answer pairs from the EgoMAN pretraining split (Sec.~\ref{sec:benchmark}). Semantic questions requiring natural language answers are supervised with the standard next-token prediction loss ($\mathcal{L}_{\text{text}}$).
In contrast, queries requiring numeric outputs (\eg, timestamps, 6DoF location), such as spatial reasoning queries, append a special token \texttt{<HOI\_Query>} to the question end, instructing the model to decode trajectory tokens. For these queries, in addition to the language modeling loss that supervises the special token as text ($\mathcal{L}_{\text{text}}$), we supervise the \texttt{<ACT>} token with an action-semantic loss ($\mathcal{L}_{\text{act}}$) and the waypoint tokens with a dedicated waypoint loss ($\mathcal{L}_{\text{wp}}$).

Specifically, we calculate the action-semantic loss by projecting the hidden state of \texttt{<ACT>} to a semantic embedding and contrast against a CLIP-encoded~\cite{radford2021learning} GT embedding. To stabilize training under varying batch sizes caused by the flexible mix of query types, with questions requiring an \texttt{<ACT>} answer varying in proportion across batches, we adaptively use cosine similarity or InfoNCE~\cite{infonce}:
\[
\small
\mathcal{L}_{\text{act}} =
\begin{cases}
1 - \tfrac{1}{K}\sum_{i=1}^{K} \text{sim}(z_i, z_i^{+}), & K < \kappa, \\[8pt]
-\tfrac{1}{K} \sum_{i=1}^{K} \log \frac{\exp(\text{sim}(z_i, z_i^{+})/\tau)}
{\sum_{j=1}^{K} \exp(\text{sim}(z_i, z_j^{+})/\tau)}, & K \ge \kappa .
\end{cases}
\]
\noindent where \(z_i\) and \(z_i^{+}\) denote normalized predicted and GT embeddings, \(\text{sim}(\cdot)\) is cosine similarity, and \(\tau\) is a learnable temperature parameter. When the number of valid training samples \(K\) falls below a threshold \(\kappa\), we apply a cosine similarity loss to avoid unstable contrastive updates; otherwise, we use an InfoNCE-style contrastive loss.

For waypoint learning, each waypoint token is supervised with Huber losses weighted by Gaussian time windows:
\[
\mathcal{L}_{\text{wp}} =
\lambda_t \mathcal{L}_{\text{time}} +
\lambda_{3D} \mathcal{L}_{3D} +
\lambda_{2D} \mathcal{L}_{2D} +
\lambda_{r} \mathcal{L}_{\text{rot6D}} +
\lambda_{\text{geo}} \mathcal{L}_{\text{geo}}.
\]
We use the continuous 6D rotation parameterization~\cite{zhou2019continuity} with a geodesic rotation loss ($\mathcal{L}{\text{geo}}$), and compute the 2D loss ($\mathcal{L}{2D}$) by projecting predicted 3D positions into the input image frame. Only visible waypoints are supervised to avoid ambiguity.

The complete reasoning pre-training loss is:
\[
\mathcal{L}_{\text{total}} =
\mathcal{L}_{\text{text}} +
\lambda_{\text{wp}} \mathcal{L}_{\text{wp}} +
\lambda_{\text{act}} \mathcal{L}_{\text{act}},
\]
where the $\lambda$ terms weight each loss component.

\subsection{Motion Expert}
\label{sec:motion_pre-training}
Given the trajectory tokens from the Reasoning Module, the Motion Expert predicts high-frequency 6DoF wrist trajectories by modeling fine-grained hand dynamics. The Motion Expert is an encoder–decoder transformer using Flow Matching (FM)~\cite{lipman2023flow} conditioned on past wrist motion, intent semantics, low-level visual features, and stage-aware waypoints. FM learns a conditional velocity field that yields smooth, probabilistic trajectories, with the three waypoint tokens providing structural guidance.

As shown in Fig. \ref{fig:main} (b), the encoder organizes all inputs into a unified sequence.
Motion-related tokens lie on a unified temporal axis: $H$ past wrist motion points occupy steps $1$--$H$, waypoint tokens are placed at predicted timestamps offset by $H$, and future queries span $H{+}1$--$H{+}T$.
These temporal tokens receive positional IDs based on their timestamps. In parallel, intent semantics and DINOv3~\cite{simeoni2025dinov3} visual features are added as non-temporal context tokens. The decoder then generates the $T$ future 6DoF trajectory points by attending to this complete encoded context.

We follow the standard FM: a noisy sample \(x_0\) is interpolated with the ground truth \(x_1\), and the supervision target is $\hat{v} = x_1 - x_0$. The loss is a mean squared error over 3D positions and 6D rotations:
\[
\mathcal{L}_{\text{FM}} = \left\| \hat{v} - (x_1 - x_0) \right\|_2^2.
\]

At test time, we sample an initial random trajectory \(x_0\) and integrate the velocity field over \(N\) steps:
\[
x_{k+1} = x_k + \Delta t \cdot \hat{v}(x_k, t_k), \quad \Delta t = \tfrac{1}{N},
\]
to obtain future wrist trajectories.

\paragraph{Motion Pre-Training.}
We found joint training of the Reasoning Module and the Motion Expert is unstable due to mismatched learning objectives. 
To address this, we pre-trained the FM model separately on the EgoMAN finetuning split (Sec.~\ref{sec:benchmark}), using GT waypoints and action phrase semantics as conditions. This provides strong low-level motion prior that stabilizes joint training with Reasoning Module.

\begin{table*}[t]
\centering
\setlength{\tabcolsep}{4pt}
\renewcommand{\arraystretch}{0.9}
\vspace{-0.2in}
\begin{tabular}{l|ccc|ccc|ccc|ccc|c}
\toprule
\multirow{2}{*}{\textbf{Method}} 
& \multicolumn{3}{c|}{\textbf{ADE (m)} $\downarrow$} 
& \multicolumn{3}{c|}{\textbf{FDE (m)} $\downarrow$} 
& \multicolumn{3}{c|}{\textbf{DTW (m)} $\downarrow$} 
& \multicolumn{3}{c|}{\textbf{Rot ($^\circ$)} $\downarrow$} 
& \multirow{2}{*}{\textbf{Dataset}} \\
\cmidrule(lr){2-4} \cmidrule(lr){5-7} \cmidrule(lr){8-10} \cmidrule(l){11-13}
& K{=}1 & K{=}5 & K{=}10 
& K{=}1 & K{=}5 & K{=}10
& K{=}1 & K{=}5 & K{=}10
& K{=}1 & K{=}5 & K{=}10
& \\
\midrule
USST*~\cite{BaoUSST_ICCV23}
&  0.233  &  0.233 &  0.233
& 0.394 & 0.394 & 0.394
& 0.220  & 0.220 & 0.220
& 46.98 & 46.98 & 46.98
& \multirow{6}{*}{\shortstack{EgoMAN\\Unseen}} \\
MMTwin*~\cite{ma2025mmtwin}
& 0.213 & 0.208 & 0.206
& 0.261 & 0.257 & 0.256
& 0.211 & 0.206 & 0.204
& 49.53 & 49.12 & 48.98
& \\
HandsOnVLM*~\cite{bao2025handsonvlm}
& 0.176 & 0.172 & 0.171
& 0.232 & 0.228 & 0.228
& 0.166 & 0.162 & 0.161
& \underline{35.49} & 35.29 & 35.22
& \\
FM-base
& 0.188 & 0.166 & 0.160
& 0.265 & 0.236 & 0.229
& 0.171 & 0.150 & 0.144
& 37.92 & 37.27 & 37.00
& \\
\cellcolor{mygray}EgoMAN-ACT
&\cellcolor{mygray}\underline{0.162} & \cellcolor{mygray}\underline{0.146} & \cellcolor{mygray}\underline{0.141}
& \cellcolor{mygray}\underline{0.225} & \cellcolor{mygray}\underline{0.210} & \cellcolor{mygray}\underline{0.204}
&\cellcolor{mygray}\underline{0.148} & \cellcolor{mygray}\underline{0.132} &\cellcolor{mygray}\underline{0.127}
&\cellcolor{mygray}36.24 & \cellcolor{mygray}\underline{35.28} & \cellcolor{mygray}\underline{35.03}
& \\
\cellcolor{mygray}EgoMAN (Ours)
& \cellcolor{mygray}\textbf{0.151} & \cellcolor{mygray}\textbf{0.130} & \cellcolor{mygray}\textbf{0.124}
& \cellcolor{mygray}\textbf{0.206} & \cellcolor{mygray}\textbf{0.186} & \cellcolor{mygray}\textbf{0.179}
& \cellcolor{mygray}\textbf{0.137} & \cellcolor{mygray}\textbf{0.117} & \cellcolor{mygray}\textbf{0.111}
& \cellcolor{mygray}\textbf{33.88} & \cellcolor{mygray}\textbf{33.00} &\cellcolor{mygray}\textbf{32.75}
& \\
\midrule
USST*~\cite{BaoUSST_ICCV23}
& 0.245 & 0.245 & 0.245
& 0.409 & 0.409 & 0.409
& 0.226 & 0.226 & 0.226
&  55.80 &  55.80 &  55.80
& \multirow{6}{*}{\shortstack{HOT3D\\OOD}} \\
MMTwin*~\cite{ma2025mmtwin}
& 0.214 & 0.210 & 0.209
& 0.263 & 0.260 & 0.259
& 0.212 & 0.208 & 0.207
& 44.75 & 44.46 & 44.37
& \\
HandsOnVLM*~\cite{bao2025handsonvlm}
& 0.197 & 0.194 & 0.194
& 0.266 & 0.262 & 0.262
& 0.191 & 0.188 & 0.186
& \underline{38.29} & \underline{38.18} & \underline{38.13}
& \\
FM-base
& 0.176 & 0.165 & 0.161
& 0.252 & 0.241 & 0.237
& 0.161 & 0.151 & 0.147
& 40.23 & 39.64 & 39.47
& \\

\cellcolor{mygray}EgoMAN-ACT
& \cellcolor{mygray}\underline{0.172} & \cellcolor{mygray}\underline{0.158} & \cellcolor{mygray}\underline{0.153}
& \cellcolor{mygray}\underline{0.247} & \cellcolor{mygray}\underline{0.232} &\cellcolor{mygray}\underline{0.228}
& \cellcolor{mygray}\underline{0.159} & \cellcolor{mygray}\underline{0.145} & \cellcolor{mygray}\underline{0.141}
& \cellcolor{mygray}39.60 & \cellcolor{mygray}38.68 & \cellcolor{mygray}38.42
& \\
\cellcolor{mygray}EgoMAN (Ours)
& \cellcolor{mygray}\textbf{0.166} & \cellcolor{mygray}\textbf{0.147} & \cellcolor{mygray}\textbf{0.141}
& \cellcolor{mygray}\textbf{0.246} &\cellcolor{mygray}\textbf{0.224} &\cellcolor{mygray}\textbf{0.217}
& \cellcolor{mygray}\textbf{0.155} & \cellcolor{mygray}\textbf{0.137} &\cellcolor{mygray}\textbf{0.130}
&\cellcolor{mygray}\textbf{36.11} &\cellcolor{mygray}\textbf{35.40} &\cellcolor{mygray}\textbf{35.09}
& \\
\bottomrule
\end{tabular}
\vspace{-0.1in}
\caption{\textbf{Comparison of 6DoF hand trajectory prediction on EgoMAN-Unseen and HOT3D-OOD.} Lower is better. Best values are \textbf{bold}, second-best are \underline{underlined}. Our EgoMAN model outperforms the strongest external baseline (HandsOnVLM) by 27.5\% ADE on both the held-out EgoMAN-Unseen test split and the out-of-distribution HOT3D-OOD dataset.}

\label{tab:merged_egoman_hot3d}
\vspace{-0.7ex}
\end{table*}

\subsection{Reasoning to Motion}
\label{sec:joint_training}
Once both the Reasoning Module and Motion Expert are pretrained, we jointly train them to connect high-level reasoning with low-level motion generation.
A key challenge in this stage is the distribution mismatch. The Reasoning Module was pretrained to predict tokens based on ground-truth, while the Motion Expert was pretrained to consume ground-truth waypoints and action phrases. At inference, however, the Motion Expert must consume the predicted (and potentially noisy) tokens from the Reasoning Module.
To bridge this gap, we align the two components through joint training on the Trajectory-Token Interface.

In the full EgoMAN model, the Reasoning Module is prompted with QA-style input, e.g., \textit{Given the past wrist motion: \{past\_motion\}. Where will the hands move to \{intent\}?}\texttt{<HOI\_QUERY>}, and produces the structured trajectory token sequence \texttt{<ACT><START><CONTACT><END>}. These tokens are then decoded into Motion Expert inputs: \texttt{<ACT>} yields an action-semantic embedding that replaces the ground-truth phrase embedding, while \texttt{<START>}, \texttt{<CONTACT>}, and \texttt{<END>} decode into 6DoF waypoints and timestamps (i.e., their positional encodings), replacing ground-truth waypoints.
To align the reasoning and motion components, we jointly train them on the EgoMAN finetuning dataset using two objectives: (1) a next-token prediction loss $\mathcal{L}_{\text{text}}$ over the trajectory-token sequence, and (2) the Flow Matching loss $\mathcal{L}_{\text{FM}}$ on the trajectories generated by the Motion Expert, as in Sec.~\ref{sec:motion_pre-training}. This unified setup enables efficient intent reasoning and produces physically consistent 6DoF trajectories aligned with the intent semantics.
\section{Experiments}
\label{sec:experiments}

\begin{figure*}[t]
\centering

\begin{minipage}[t]{0.6\textwidth}\vspace{0pt}
\centering
\small
\setlength{\tabcolsep}{2pt}
\renewcommand{\arraystretch}{0.9}
\begin{tabular}{l|ccc|cc|cc}
\toprule
\multirow{2}{*}{\textbf{Method}} & \multirow{2}{*}{\textbf{WP}} & \multirow{2}{*}{\textbf{Detect}} & \multirow{2}{*}{\textbf{FPS} $\uparrow$} &
\multicolumn{2}{c|}{\textbf{EgoMAN}} &
\multicolumn{2}{c}{\textbf{HOT3D}} \\
 &  &  &  & \textbf{Contact} $\downarrow$ & \textbf{Traj} $\downarrow$ &
\textbf{Contact} $\downarrow$ & \textbf{Traj} $\downarrow$ \\
\midrule

HAMSTER*~\cite{li2025hamster}   & 2D-text   &\xmark  &0.17  &  0.342 & 0.297 & 0.236 & 0.219 \\
VRB*~\cite{vrb}       & 3D    &\cmark &0.03 & 0.300 &0.271  & 0.216 & 0.224 \\
VidBot~\cite{chen2025vidbot}    & 3D     &\cmark &0.04 & 0.290 & 0.269 & 0.190 & 0.147 \\
\cellcolor{mygray}{EgoMAN-WP}     &\cellcolor{mygray}3D & \cellcolor{mygray}\xmark & \cellcolor{mygray}\textbf{3.45} & \cellcolor{mygray}\textbf{0.192} &\cellcolor{mygray}\textbf{0.127} & \cellcolor{mygray}\textbf{0.188} & \cellcolor{mygray}\textbf{0.110} \\
\bottomrule
\end{tabular}
\vspace{-0.1in}
\captionof{table}{\textbf{Waypoint prediction results.}
Lower is better for \emph{Contact} and \emph{Traj}; higher is better for \emph{FPS} (averaged over 50 samples on an NVIDIA PG509-210, 80GB). 
\emph{EgoMAN-WP} achieves the best accuracy, improving Contact by 33.8\% and Traj by 52.8\% on EgoMAN-Unseen, and runs orders of magnitude faster at 3.45 FPS.}

\label{tab:wp}
\end{minipage}
\hfill
\begin{minipage}[t]{0.38\textwidth}\vspace{0pt}
\centering
\small

\setlength{\tabcolsep}{2pt}
\renewcommand{\arraystretch}{0.9}
\scalebox{0.94}{
\begin{tabular}{cc|c|cccc}
\toprule
\multicolumn{2}{c|}{\textbf{Pretrain}} 
& \multirow{2}{*}{\textbf{WP}}
& \multirow{2}{*}{\textbf{ADE} $\downarrow$}
& \multirow{2}{*}{\textbf{FDE} $\downarrow$}
& \multirow{2}{*}{\textbf{DTW} $\downarrow$}
& \multirow{2}{*}{\textbf{Rot} $\downarrow$} \\
\textbf{Reason} & \textbf{FM} &  &  &  &  & \\
\midrule

\xmark & \xmark & \xmark
& 0.273 & 0.308 & 0.260 & 51.79 \\

\cmark & \xmark & 6DoF
& 0.215 & 0.255 & 0.198 & 43.03 \\

\xmark & \cmark & \xmark
& 0.162 & 0.225 & 0.148 & 36.24 \\

\cmark & \cmark & \xmark
& 0.161 & 0.224 & 0.147 & 35.90 \\

\cmark & \cmark & Emb
& \textbf{0.150}
& \underline{0.210}
& \underline{0.138}
& \underline{34.02} \\

\cmark & \cmark & 6DoF
& \underline{0.151}
& \textbf{0.206}
& \textbf{0.137}
& \textbf{33.88} \\
\bottomrule
\end{tabular}
}
\vspace{-0.1in}
\captionsetup{type=table,singlelinecheck=true}
\captionof{table}{\textbf{Ablation on EgoMAN-Unseen ($K{=}1$).} Lower is better. \emph{Reason} and \emph{FM} pretraining with \emph{6DoF} waypoints yield the highest accuracy.}

 \label{tab:ablation}
\end{minipage}
\end{figure*}

\begin{figure*}[t!]
  \centering
  \includegraphics[width=.9\linewidth]{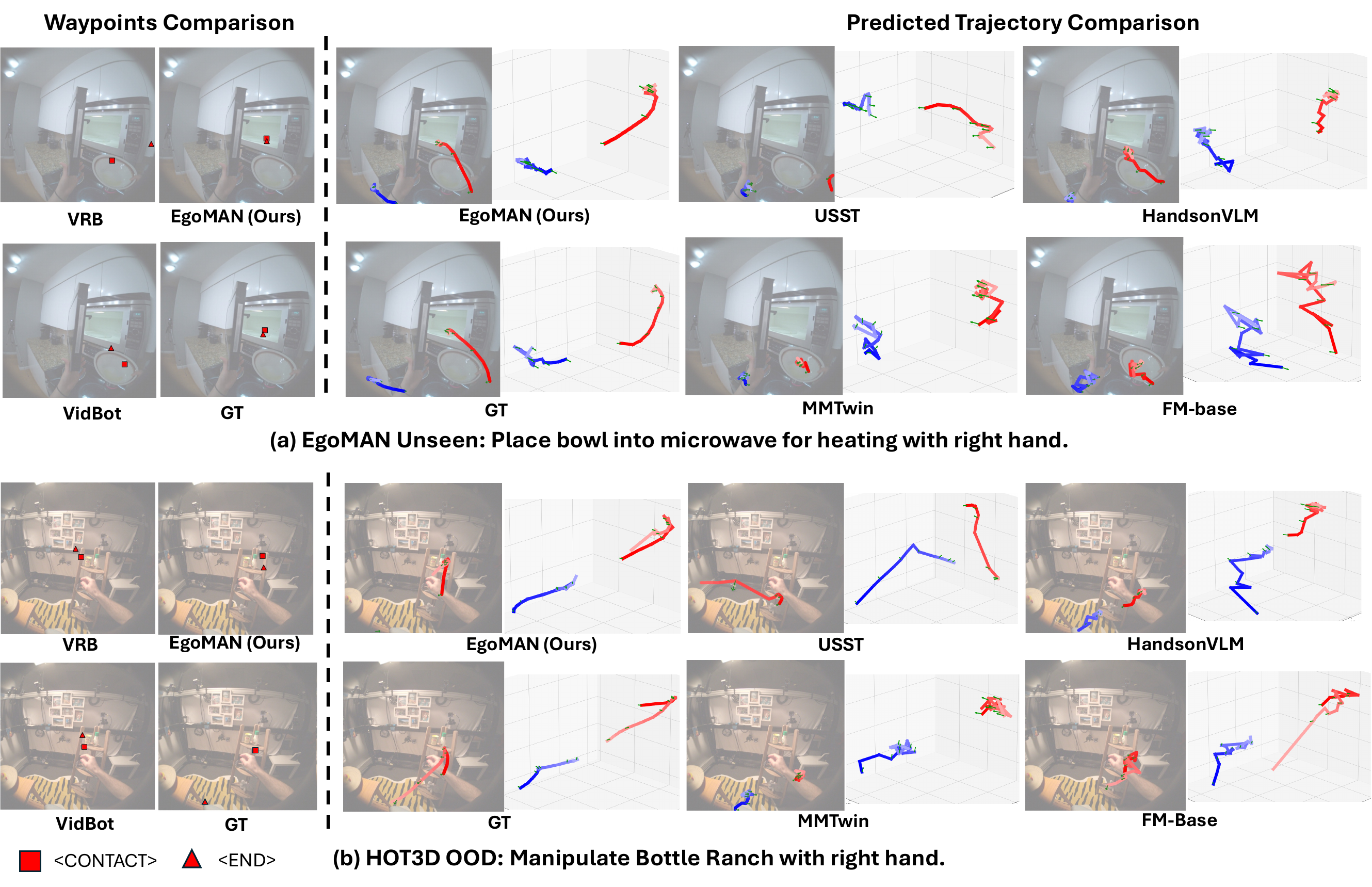}
    \vspace{-1em}
  \caption{
\textbf{Qualitative comparisons on EgoMAN-Bench.} We visualize best-of-$K{=}10$ predictions for waypoints and full trajectories. Left: \texttt{<CONTACT>} and \texttt{<END>} waypoint predictions compared with \emph{VRB*} and \emph{VidBot}. Right: 3D hand trajectory forecasts and 2D projections compared with prior baselines. Our \emph{EgoMAN} model produces the smoothest and closest results to ground truth.
}
  \vspace{-1em}
  \label{fig:qualitative_compbest_egoman}
\end{figure*}

We evaluate the \emph{EgoMAN} model thoroughly on EgoMAN-Bench to answer three core questions:
(1) \emph{Does the reasoning-to-motion pipeline improve long-horizon 6DoF prediction over state-of-the-art baselines?}
(2) \emph{How effectively does the Reasoning Module generate accurate and reliable waypoints for intent-aligned spatial prediction?}
(3) \emph{How do the progressive training strategy and the trajectory-token interface contribute to overall performance?}
We further provide qualitative results showing diverse generalization and controllable intent-conditioned motion.

\subsection{Evaluation Setting and Metrics}
\noindent\textbf{Trajectory Metrics.}
We evaluate all methods using standard hand–trajectory forecasting metrics, including \textbf{Average Displacement Error (ADE)}, \textbf{Final Displacement Error (FDE)}, and \textbf{Dynamic Time Warping (DTW)}, all reported in meters, as well as \textbf{Angular Rotation Error (Rot)} in degrees. To assess stochastic generative prediction, each model samples $K{=}1/5/10$ trajectories per query. Unless otherwise specified, all results are reported as \textbf{best-of-$K$}, which selects the trajectory with minimum error to the ground truth. 

\noindent\textbf{Waypoint (WP) Metrics.} 
We evaluate the \texttt{<CONTACT>} and \texttt{<END>} waypoints predicted by our VLM. We report two metrics in meters to quantify the localization accuracy of key intent states:
\textbf{Contact Distance (Contact):} The Euclidean distance between the predicted and ground-truth wrist locations at the approach-completion timestamp.
and \textbf{Trajectory-Warp Distance (Traj):} The average Euclidean distance from each predicted waypoint to its nearest point on the GT trajectory.

\subsection{Baselines}
\noindent\textbf{Hand Trajectory Predictor Baselines.}
We compare against five trajectory baselines. Baselines marked with (*) are adapted for fair comparison by matching the \emph{EgoMAN} setting: using a single RGB image, an intent text embedding, and past motion as inputs to predict up to 5-second 6DoF bi-hand trajectories, with metrics computed over the ground-truth duration.
1) \textbf{USST*}~\cite{BaoUSST_ICCV23}: an uncertainty-aware state–space transformer for egocentric 3D hand trajectory forecasting;  
2) \textbf{MMTwin*}~\cite{ma2025mmtwin}: a model using twin diffusion experts and a Mamba–Transformer backbone for joint egomotion and hand motion prediction; 
3) \textbf{HandsOnVLM*}~\cite{bao2025handsonvlm}: a VLM that predicts 2D trajectories via dialogue, which we adapt to 6DoF poses using a Conditional Variational Autoencoder  (CVAE)~\cite{cvae} head from 50 predicted special hand tokens; 
We also include two ablations from our own pipeline:  
4)~\textbf{FM-Base}: our Flow Matching Motion Expert conditioned on image, intent, and past motion, but without VLM reasoning; and
5)~\textbf{EgoMAN-ACT}: our variant that removes reasoning pre-training and waypoint supervision, conditioning the Motion Expert only on a VLM-learned \texttt{<ACT>} token.

\noindent\textbf{Affordance-driven Baselines.}
We evaluate three affordance-based methods: \textbf{HAMSTER*}~\cite{li2025hamster}, \textbf{VRB*}~\cite{vrb}, and \textbf{VidBot}~\cite{chen2025vidbot}, each adapted to our waypoint prediction setting for a fair comparison. All methods take the same RGB image, Metric3D depth~\cite{hu2024metric3dv2}, and verb–object text as input, and predict contact and goal points aligned with EgoMAN’s \texttt{<CONTACT>} and \texttt{<END>} waypoints. Aria fisheye images are rectified to pinhole views using device calibration. \textit{VRB*} and \textit{HAMSTER*} produce 2D affordance points that we unproject to 3D, and for \textit{HAMSTER*} we treat the first and last predicted points as contact and goal. \textit{VidBot} and \textit{VRB*} return object-conditioned affordance candidates; when multiple candidates appear, we select the one closest to the target object. Since these models output affordance points rather than wrist poses, we approximate wrist locations by choosing the predicted point closest to the GT wrist within 5 cm. 

\subsection{Results}
\paragraph{Trajectory Evaluation.}
As shown in Table~\ref{tab:merged_egoman_hot3d}, the full \emph{EgoMAN} model achieves the lowest ADE, FDE, DTW, and rotation errors for all sampling budgets $K \in {1,5,10}$. It outperforms the strongest external baseline \emph{HandsOnVLM*} by a large margin, reducing ADE at $K{=}10$ by 27.5\% on the held-out EgoMAN-Unseen split and achieving a similar 27.3\% reduction on the out of distribution HOT3D-OOD dataset, demonstrating strong generalization across domains.

\noindent\emph{FM-base} already outperforms state-space and Mamba-based predictors (\emph{USST*}, \emph{MMTwin*}) on both test splits, showing that Flow Matching–based encoder–decoder modeling provides a stronger foundation for long-horizon 6DoF motion forecasting. Incorporating vision--language supervision further improves accuracy. \emph{HandsOnVLM*} leverages text guidance, but its CVAE decoder predicts trajectories from 50 VLM-produced hand tokens learned from noisy egocentric data, yielding higher 6DoF errors and showing little improvement even as the sample count $K$ increases. 

\noindent\textbf{Waypoints Evaluation.}  
Table~\ref{tab:wp} shows that \emph{EgoMAN-WP}, which directly regresses 3D \texttt{<CONTACT>} and \texttt{<END>} wrist positions from the Reasoning Module, achieves the best performance on both EgoMAN-Unseen and HOT3D-OOD. It reduces contact error from $0.29$–$0.34$,m to $0.19$,m and lowers DTW from $0.27$–$0.30$,m to $0.13$,m on EgoMAN-Unseen, while maintaining the lowest contact error on the challenging test set HOT3D-OOD. \emph{EgoMAN-WP} is also far more efficient, running at $3.45$,FPS compared to $<0.05$,FPS for \emph{VRB*} and \emph{VidBot}, which require heavy detection and 3D post-processing. By predicting only four structured and compact trajectory tokens, our \emph{EgoMAN} model gains both speed and robustness.

\subsection{Ablation Study}
\label{sec:main_ablation}
As shown in Table~\ref{tab:ablation}, we ablate different components of our method. For pretraining, we toggle \textbf{Reasoning Pretrain} (Reason) and \textbf{Flow-Matching Pretrain} (\emph{FM}). For \textbf{Waypoint} (\emph{WP}) choices, we compare using no waypoints ($\mathcal{X}$), our explicit 6DoF waypoints (\emph{6DoF}), and decoder’s final hidden state as implicit embeddings (\emph{Emb}). 

\noindent Starting from a model without any pretraining or waypoints, adding only \emph{FM} pretraining (\xmark/\cmark/\xmark) yields the largest single gain across all metrics, showing the importance of a motion-aware initialization. \emph{Reason} pretraining alone (\cmark/\xmark/\emph{6DoF}) also provides substantial improvements, but remains weaker than \emph{FM} pretraining. Combining both pretraining signals (\cmark/\cmark/\xmark) further reduces error, indicating complementary benefits. Finally, with \emph{Reason} and \emph{FM} pretraining, adding the waypoint interface yields the strongest overall performance: the implicit waypoint variant \emph{Emb} achieves the lowest ADE and closely approaches the full model with explicit \emph{6DoF} waypoint design, which delivers the lowest FDE, DTW, and rotation errors. Please see more detailed analysis of the ablation results in appendix.

\begin{figure}[t]
  \centering
  \includegraphics[width=\linewidth]{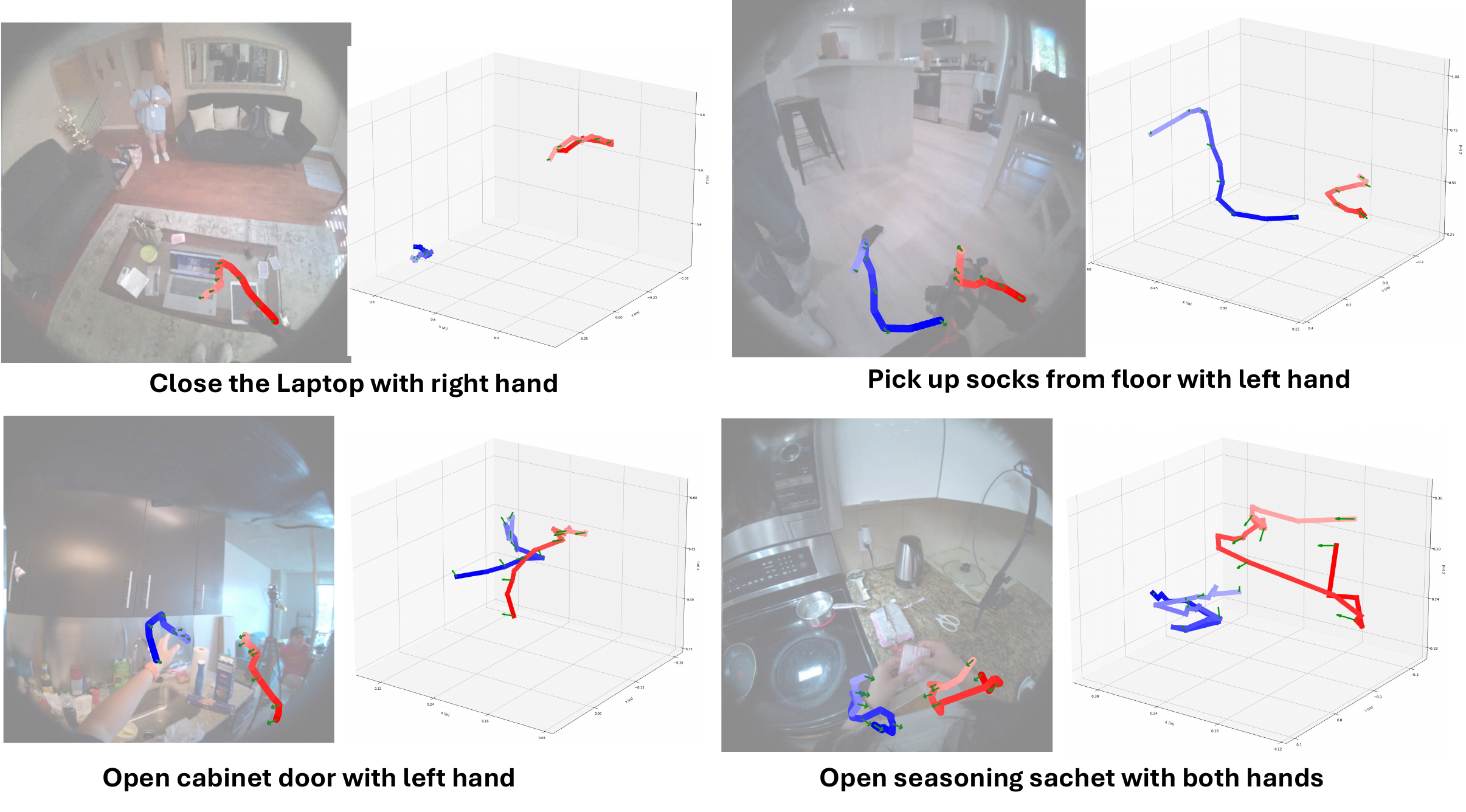}
    \captionof{figure}{
    \textbf{Qualitative results of diverse activities.}
    EgoMAN generates accurate 6DoF hand trajectories for diverse activities, aligning motion with the intent description and scene spatial.
    }
    \label{fig:visdiverse}
    \vspace{-1.5em}
\end{figure}

\begin{figure}[t]
  \centering
    \includegraphics[width=\linewidth]{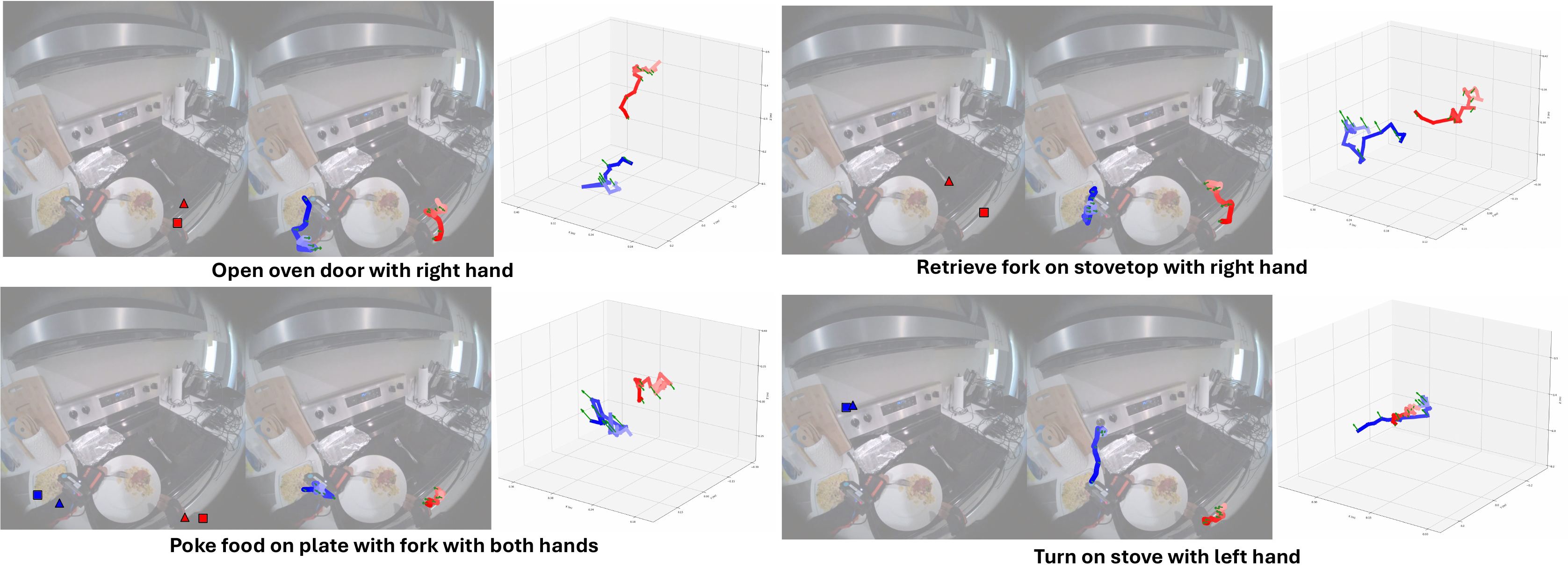}
\captionof{figure}{\textbf{Multiple intents.} With the same image and past motion, \emph{EgoMAN} model produces distinct 6DoF trajectories for different intent queries, showing controllable intent-to-motion generation.} 
    \label{fig:duotextvislarge}
    \vspace{-1em}
\end{figure}

\subsection{Qualitative Analysis}
We visualize qualitative best-of-$K{=}10$ forecasts in Figure~\ref{fig:qualitative_compbest_egoman} on EgoMAN-Bench (EgoMAN-Unseen and HOT3D-OOD).
On the left of Figure~\ref{fig:qualitative_compbest_egoman}, we compare waypoints from affordance baselines (\emph{VRB*}, \emph{VidBot}) with those from our \emph{EgoMAN} model. Our predicted contact and end points align closely with the GT wrist positions, while affordance methods often miss the target surface due to detection errors or collapse toward the hand instead of the intended goal region.
On Figure~\ref{fig:qualitative_compbest_egoman} right, we compare full 6DoF trajectories against trajectory baselines (\emph{USST}, \emph{HandsOnVLM}, \emph{MMTwin}, \emph{FM-base}). \emph{EgoMAN} generates smoother and more accurate motions that reach the target and complete the manipulation with correct wrist orientation, while baselines often underreach, overshoot, or drift in cluttered scenes or under unfamiliar objects and intent descriptions.

\noindent Figure~\ref{fig:visdiverse} further illustrates diverse activities such as closing a laptop, picking up socks, opening a cabinet, and opening a seasoning sachet. Across these scenarios, \emph{EgoMAN} model consistently produces trajectories that follow the verb phrase and scene spatial contexts, demonstrating that the reasoning-to-motion pipeline generalizes well to a wide range of real-world hand–object interactions.

\noindent Finally, Figure~\ref{fig:duotextvislarge} illustrates intent-conditioned motion generation under the same visual and motion context.
Given different intent queries, \emph{EgoMAN} model reasons about the action, predicts distinct waypoints, and produces correspondingly different yet valid 6DoF trajectories (\eg opening the oven, retrieving a fork, poking food, turning on the stove).
This controllable intent-to-motion mapping enables flexible generation of diverse hand trajectories, which can support robot learning and data augmentation.
\section{Conclusion}
We introduced the \textbf{EgoMAN dataset}, a large-scale egocentric benchmark for interaction stage–aware 6DoF hand trajectory prediction, featuring structured QA pairs that capture semantic, spatial, and motion reasoning. We also presented the \textbf{EgoMAN model}, a modular reasoning-to-motion framework that aligns high-level intent with physically grounded 6DoF trajectories through a trajectory-token interface and progressive training. Experiments show strong gains over both motion-only and VLM baselines: Flow Matching yields smoother and more stable trajectories, VLM-driven reasoning improves semantic alignment and generalization to novel scenes and intents, and the trajectory-token interface enables efficient inference, bridging intent-conditioned stage–aware reasoning with precise low-level motion generation. Overall, \textbf{EgoMAN} offers a practical step toward in-context action prediction, supporting applications in robot manipulation, language-conditioned motion synthesis, and intent-aware assistive systems.

\section*{Acknowledgment}
We thank Lingni Ma for valuable discussions and support related to the Nymeria dataset. We also thank Hanzhi Chen for assisting with adapting the affordance model for our waypoint evaluation protocol. 
We are grateful to the teams behind Project Aria, EgoExo4D, and HOT3D for releasing the foundational datasets that enabled this research.

{
    \small
    \bibliographystyle{ieeenat_fullname}
    \bibliography{main}
}
\clearpage
\newpage
\clearpage
\appendix

\section{Summary of Appendix}
\noindent In this appendix, we provide:
\begin{enumerate}
    \item A video demonstration of our system, including representative interaction cases (Sec.~\ref{sec:case_video}).
    \item Implementation details of the EgoMAN model pipeline and our progressive training strategy (Sec.~\ref{sec:implementation}).
    \item Extended analysis of ablation results presented in the main paper (Sec.~\ref{sec:suppl_ablations}).
    \item Evaluation of semantic alignment between predicted hand trajectories and action verbs, with comparisons to baselines (Sec.~\ref{sec:motion2text}).
    \item Comparison across different parameter scales of the Reasoning Module, including EgoMAN-QA accuracy and trajectory prediction performance (Sec.~\ref{sec:reason_scale}).
    \item Representative prompt examples used in data annotation (Sec.~\ref{sec:prompt}).
    \item Limitations and future work (Sec.~\ref{sec:limitations}).
\end{enumerate}

\section{Video}
\label{sec:case_video}
Our
\href{https://egoman-project.github.io/static/videos/egoman_suppl_video.mp4}{\texttt{video}}
 provides a visual overview of the core contributions of EgoMAN, covering the dataset, the model architecture, and qualitative demonstrations. The video first introduces the EgoMAN dataset and highlights the full EgoMAN pipeline, consisting of the Reasoning Module, the Motion Expert, and the end-to-end 6-DoF trajectory generation flow (bridging reasoning to motion through the Trajectory-Token Interface).

In the dataset overview segment, we present diverse examples of hand--object interactions corresponding to the twelve most frequent verbs in EgoMAN (e.g., \emph{Grasp}, \emph{open}, \emph{place}, \emph{pour}, \emph{stir}). The showcased clips span multiple sources such as EgoExo4D~\cite{egoexo4d}, Nymeria~\cite{nymeria}, and HOT3D-Aria~\cite{hot3d}, illustrating that interactions occur in realistic everyday scenes with natural noise, clutter, and challenging viewpoints. These examples demonstrate the dataset’s scale, variability, and difficulty, motivating the need for robust intention-conditioned 3D hand trajectory modeling.

The qualitative results section highlights EgoMAN’s ability to generate intention-guided trajectories. We first show dozens of representative cases across various scenarios: such as \emph{stir milk} and \emph{turn off stove} in kitchen scenes, \emph{pick up socks} and \emph{open door} in household scenes, \emph{close laptop} and \emph{grab cable} in working scenarios, and \emph{manipulate bowl} or \emph{manipulate ranch bottle} in HOT3D scenes. For each case, the video displays (1) the original input image, (2) the intention text, (3) intermediate waypoint predictions from the Reasoning Module, and (4) the final 6-DoF trajectory output. These predictions are visualized both on the static input image and overlaid on future ego-video frames to more clearly illustrate spatial accuracy and motion quality.

Across all demonstrations, EgoMAN consistently predicts accurate contact and end-point waypoints around target objects, with the generated 3D trajectories following realistic manipulation paths that match the intended semantics. While certain open-ended tasks (e.g., \emph{open door}, \emph{pick up socks}) may exhibit slight variations in final pose or timing, or minor deviations in the non-manipulating hand relative to the single ground-truth instance, the predicted trajectories for the manipulating hand remain semantically aligned with the intended goal.
These results highlight EgoMAN's capability to produce reliable, intention-driven 6-DoF hand trajectories across diverse scenes and interaction types.

We further show results of \emph{goal-directed trajectory generation}, where the same input image paired with different intention descriptions. EgoMAN model is able to predict trajectories in distinct that align with the intended goals, even in unseen environments. 

Please visit our \href{https://egoman-project.github.io}{\texttt{project website}} to check more trajectory prediction results in diverse interaction scenarios.

\section{Implementation Details}  
\label{sec:implementation}
\noindent\textbf{Reasoning Module.} The Reasoning Module is optimized in \texttt{bf16} using AdamW~\cite{Loshchilov2017DecoupledWD} with cosine learning rate decay. The vision encoder and multimodal projector are frozen. We use a base learning rate of $1{\times}10^{-5}$, a warmup ratio of $0.02$, weight decay of $0.05$, maximum gradient norm of $1.0$, and a batch size of $256$ across 8$\times$NVIDIA A100 80GB GPUs. Training runs for 2 epochs on approximately 1M EgoMAN QA samples. Images use dynamic resizing with \texttt{max\_pixels=50176} and \texttt{min\_pixels=784}.  If past motion is provided in the input question, we use the most recent 5 past points of both hands tokenized at 10\,fps; otherwise a zero-initialized motion history is used. A 4-layer MLP is used to extract features from the motion, which are then fed into Qwen2.5-VL~\cite{qwen2025qwen25technicalreport}.
The specialized waypoint decoders are lightweight ReLU MLPs with hidden dimension 768, predicting timestamp, 3D position, and 6DoF rotation. The action semantic decoder is a single-layer MLP (dim 768). When valid samples in a batch fall below \(\kappa{=}10\), we use cosine similarity loss; otherwise, we apply an InfoNCE loss~\cite{infonce}.
Loss weights are set as $\lambda_{\text{wp}}{=}0.3$ and $\lambda_{\text{act}}{=}0.1$, with internal weights $\lambda_{t}{=}1.0$, $\lambda_{3D}{=}2.0$, $\lambda_{2D}{=}0.5$, $\lambda_{r}{=}0.5$, and $\lambda_{\text{geo}}{=}0.15$. We apply Huber loss with $\beta{=}0.2$ for rot6D, $\beta_{3D}{=}0.07$, and $\beta_{2D}{=}0.02$ for location terms. The geodesic rotation loss is applied only to visible waypoints. Temporal modulation is implemented using a Gaussian time window with $\sigma_{\text{time}}{=}3.0$. 

\noindent\textbf{Motion Expert.}  
We pre-train the flow-matching (FM)~\cite{lipman2023flow} based motion decoder using approximately 17K trajectories. Inputs include DINOv3 image features, ground-truth action phrases, waypoint tokens, and past wrist motion. Motion sequences are sampled at 10\,fps with a maximum 50-step future horizon (5\,s).
The FM architecture uses a hidden dimension of 768, with 6 encoder and 6 decoder transformer layers and 8 attention heads. A sinusoidal time embedding (256-D) is mapped to FiLM-style~\cite{film}$ (\gamma, \beta)$ parameters. A 2-layer self-attention block is applied before decoding, along with modality and positional embeddings.
We train in FP32 using AdamW with a learning rate of $1{\times}10^{-4}$, weight decay of $1{\times}10^{-4}$, a cosine schedule with 5\% warmup, and a batch size of 256 on a single A100 GPU. The training objective is the sum of MSE loss on 3D positions and MSE loss on 6D rotations, with a rotation loss weight of 0.5.
At inference time, we iterate for 150 steps and retain only the predicted trajectory segment beyond the length of the ground-truth target.

\noindent\textbf{Joint Training of EgoMAN Model.}  
We initialize the Reasoning Module from the reasoning pretraining checkpoint and the motion decoder from the FM pretraining weights. The training setup largely follows the reasoning pretraining configuration, but FM components are kept in FP32. We use a learning rate of $5{\times}10^{-6}$ and a batch size of 128 cross 8×NVIDIA A100 80GB GPUs. The model is trained for 60 epochs on the same finetuning trajectory dataset used in motion pretraining. At inference time, we iterate for 150 steps and retain only the predicted trajectory segment beyond the ground-truth target length.

\section{Detailed Analysis for Ablation Study}
In this section, we provide more detailed analysis of our ablation results in main paper Sec~\ref{sec:main_ablation} and Table~\ref{tab:supp_ablation} in the appendix.

\begin{table}[t]
\centering
\setlength{\tabcolsep}{3pt}      
\renewcommand{\arraystretch}{0.98} 

\resizebox{0.95\columnwidth}{!}{%
\begin{tabular}{cc|c|cccc}
\toprule
\multicolumn{2}{c|}{\textbf{Pretrain}} 
& \multirow{2}{*}{\textbf{WP}}
& \multirow{2}{*}{\textbf{ADE} $\downarrow$}
& \multirow{2}{*}{\textbf{FDE} $\downarrow$}
& \multirow{2}{*}{\textbf{DTW} $\downarrow$}
& \multirow{2}{*}{\textbf{Rot} $\downarrow$} \\
\textbf{Reason} & \textbf{FM} &  &  &  &  & \\
\midrule

\xmark & \xmark & \xmark
& 0.273 & 0.308 & 0.260 & 51.79 \\

\cmark & \xmark & 6DoF
& 0.215 & 0.255 & 0.198 & 43.03 \\

\xmark & \cmark & \xmark
& 0.162 & 0.225 & 0.148 & 36.24 \\

\cmark & \cmark & \xmark
& 0.161 & 0.224 & 0.147 & 35.90 \\

\cmark & \cmark & Emb
& \textbf{0.150}
& \underline{0.210}
& \underline{0.138}
& \underline{34.02} \\

\cmark & \cmark & 6DoF
& \underline{0.151}
& \textbf{0.206}
& \textbf{0.137}
& \textbf{33.88} \\
\bottomrule
\end{tabular}
}

\captionsetup{type=table,singlelinecheck=true}
\captionof{table}{\textbf{Ablation on EgoMAN-Unseen ($K{=}1$).} Lower is better. \emph{Reason} and \emph{FM} pretraining with \emph{6DoF} waypoints yield the highest accuracy.}
\label{tab:supp_ablation}
\end{table}

\label{sec:suppl_ablations}
\begin{figure}[t]
  \centering
\includegraphics[width=1.0\linewidth]{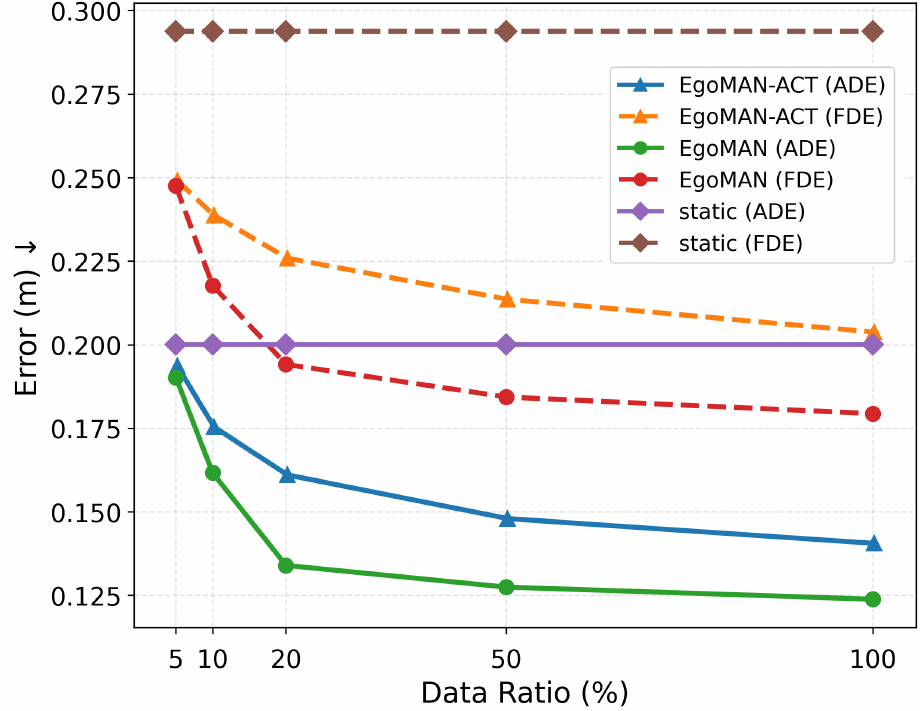}
\captionof{figure}{\textbf{Data efficiency results.}
ADE/FDE (m), best-of-10. The \textit{static} baseline repeats the last observed hand location. Without pretraining, errors of EgoMAN-ACT rise sharply under limited data, while EgoMAN maintains strong performance even at $20\%$ data, highlighting the benefit of waypoint-based Reasoning Module and pretraining.}
\label{fig:data_ratio_best}
\end{figure}

\noindent\textbf{Reasoning Pretraining.}
Removing reasoning pretraining (\emph{EgoMAN-ACT}), which also disables WP, degrades performance (ADE $0.162{\rightarrow}0.215$). This pretraining uses large-scale, noisy corpora with question-answer pairs, encoding semantic, spatial, and motion-aware priors that help disambiguate intent. Learned waypoints further align trajectories with both intent and visual context. Data-efficiency results (Fig.~\ref{fig:data_ratio_best}) support this trend: at $20\%$ of the training data, \emph{EgoMAN} (ADE $\sim0.13$\,m) remains superior to \emph{EgoMAN-ACT} (ADE $\sim0.16$\,m), and the gap persists even at full data (ADE $0.140{\rightarrow}0.125$). FDE exhibits a similar pattern.
Reasoning pretraining and waypoint conditioning reduce ambiguity early, enabling stable outputs with fewer high-quality labels.

\noindent\textbf{Trajectory-Token Interface via Waypoints (WP).}
WP provides a structured interface between the visual-language module and the 6DoF Motion Expert. Conditioning on predicted WP improves accuracy and stability (ADE $0.161{\rightarrow}0.151$, DTW $0.147{\rightarrow}0.137$, Rot $35.90^\circ{\rightarrow}33.88^\circ$). Without WP, performance drops near that of \emph{EgoMAN-ACT}, suggesting that learned WP enhance the contribution of reasoning pretraining beyond implicit semantic embeddings. Replacing 6DoF waypoints with the decoder’s final hidden state (Emb) yields only minor differences (FDE $\sim$0.4\,cm worse; other metrics nearly unchanged).

\noindent\textbf{Motion Pretraining (FM).}
Removing FM while retaining reasoning pretraining leads to clear degradation (ADE $0.150{\rightarrow}0.215$). Without FM, the model must learn both semantics and motion jointly through an implicit semantics interface, resulting in noisier long-horizon predictions and increased rotation error. Removing both FM and reasoning causes further decline (ADE $0.273$, Rot $51.79^\circ$). While reasoning pretrain and waypoints learning help mitigate this, pretraining FM on physically plausible motion first, followed by joint fine-tuning with reasoning, produces the largest improvements across all metrics.

\section{Motion-to-Text Alignment}
\label{sec:motion2text}
We evaluate semantic alignment between trajectories and action verbs by training a motion encoder to map hand trajectories to a pre-computed verb text embedding space using a CLIP-style contrastive loss~\cite{radford2021learning, chuan2022tm2t}. We report \textbf{Recall@3} (fraction of samples where the GT caption is retrieved in the top 3 over 239 verbs in the test samples) and \textbf{Fréchet Inception Distance (FID)} between predicted and ground-truth motion embeddings. To account for generative diversity, we report best-of-$K$ (K=10) retrieval results.

\begin{table}[t]
    \centering
\begin{tabular}{l|cc} 
\toprule
\textbf{Method} & \textbf{R@3} $\uparrow$ & \textbf{FID} $\downarrow$ \\
\midrule
USST*~\cite{BaoUSST_ICCV23}  & 15.0 & 0.22 \\
MMTwin*~\cite{ma2025mmtwin}         & 22.9 & 0.86 \\
HandsonVLM*~\cite{bao2025handsonvlm}   & 27.9 & 0.10  \\
FM-Base        & 39.7 & 0.05 \\
\cellcolor{mygray}EgoMAN       & \cellcolor{mygray}\bf 43.9 &\cellcolor{mygray}\bf 0.04  \\
\bottomrule
\end{tabular}
\vspace{-0.05in}
\captionof{table}{\textbf{Motion-to-Verb Text Retrieval.} Train one encoder; evaluate verb text-motion relevance over 239 verb candidates.}
\label{tab:motiontext}
\end{table}

\begin{table*}[htb]
\centering
\setlength{\tabcolsep}{4pt}
\begin{tabular}{cc|ccc|cc|ccc}
\toprule
\multirow{2}{*}{Model} & \multirow{2}{*}{Params} &
\multicolumn{3}{c|}{Spatial Reasoning (Waypoints)} &
\multicolumn{2}{c|}{Semantic Embedding} &
\multicolumn{3}{c}{Semantic Text QA} \\
& &
Loc $\downarrow$ & Time $\downarrow$ & Rot$^\circ\downarrow$ &
R@3 (\%) $\uparrow$ & Pearson $\uparrow$ &
BERT $\uparrow$ & BLEU $\uparrow$ & ROUGE $\uparrow$ \\
\midrule
Qwen2.5-VL & 3B 
& 0.229 & 0.483 & 41.36
& \underline{6.26} & \underline{0.239}
& 0.914 & 0.155 & 0.469 \\

Qwen2.5-VL & 7B 
& \underline{0.225} & \bf 0.474 & {41.27}
& \textbf{11.08} & \textbf{0.256}
& 0.916 & 0.165 & 0.481 \\
\midrule
Qwen3-VL & 2B 
& 0.244 & 0.495 & 41.88
& 1.62 & 0.107
& \textbf{0.919} & \underline{0.171} & 0.487 \\

Qwen3-VL & 4B 
& \textbf{0.223} & {0.481} & \textbf{40.89}
& 1.69 & {0.134}
& \textbf{0.919} & \underline{0.171} & \textbf{0.494} \\

Qwen3-VL & 8B 
& 0.228 & \underline{0.477} & \underline{41.24}
& {5.66} & 0.205
&  \underline{0.917} & \bf 0.172 & \underline{0.488} \\
\bottomrule
\end{tabular}
\caption{\textbf{Effect of model scale on spatial reasoning, semantic alignment, and text QA on EgoMAN Unseen benchmark.}
We evaluate (i) waypoint spatial reasoning via 3D location, time, and rotation errors,
(ii) semantic embedding alignment using R@3 (computed over 2,844 GT action-embedding candidates) and mean Pearson correlation,
and (iii) semantic text QA using BERTScore, BLEU, and ROUGE.
Best values are \textbf{bolded}; second-best are \underline{underlined}. Spatial reasoning performance saturates early, and models larger than 2B/3B provide consistently stronger performance. Semantic alignment benefits from larger models, with Qwen2.5-VL outperforming Qwen3-VL, while text QA remains relatively stable across scales, with Qwen3-VL slightly outperforming Qwen2.5-VL.
}
\label{tab:reasoning_scale_qa}
\end{table*}

Table~\ref{tab:motiontext} shows that \emph{EgoMAN} achieves the strongest semantic alignment between motion and verb phrases, with the highest R@3 ($43.9\%$) and lowest FID ($0.04$). Generative baselines such as \emph{USST} and \emph{MMTwin} produce smooth trajectories but exhibit weaker text alignment, while \emph{HandsOnVLM} benefits from language conditioning yet suffers from noisy CVAE decoding. \emph{FM-Base} already improves verb specificity, indicating that Flow Matching promotes more structured motion. Adding VLM reasoning and waypoint constraints in \emph{EgoMAN} further reduces ambiguity, tightening the trajectory–verb correspondence and producing a motion embedding distribution closer to ground truth.

\section{Reasoning Module Scale Analysis}
\label{sec:reason_scale}

In this section, we analyze how scaling the Reasoning Module affects both high-level semantic reasoning and downstream motion prediction. We evaluate multiple model sizes from Qwen2.5-VL and Qwen3-VL families, using identical training data and identical Trajectory-Token Interface settings. Our analysis focuses on two components: (i) EgoMAN QA: measuring semantic and spatial reasoning, and (ii) trajectory prediction on EgoMAN-Bench: measuring the effect of Reasoning Module scale on 6-DoF hand trajectory generation.

\subsection{EgoMAN QA}
\label{sec:reason_scale_qa}

\textbf{Evaluation Metrics.}
We evaluate three complementary aspects of reasoning quality:

\begin{itemize}[leftmargin=5mm]
    \item \textbf{Waypoint Spatial Reasoning}: 
    We evaluate 3D waypoint accuracy for  \texttt{<CONTACT>} and \texttt{<END>} using three metrics. 
    \textbf{Location error (Loc)} is the Euclidean distance (in meters) between the predicted and ground-truth waypoints' positions. 
    \textbf{Time error (Time)} measures the temporal accuracy of the predicted interaction stages. 
Since \texttt{<START>} (approach onset) is always aligned to time~0, we compute the L1 difference (in seconds) only for the predicted \texttt{<CONTACT>} (manipulation onset) and \texttt{<END>} (manipulation completion) timestamps.

    \textbf{Rotation error (Rot)} is the geodesic distance (in degrees) between predicted and ground-truth wrist orientations, computed from the relative rotation matrix. 
    These metrics quantify spatial, temporal, and rotational grounding of interaction-stage waypoints.

    \item \textbf{Semantic Embedding Alignment}: we compute \textbf{Recall@3 (R@3)} between predicted and 2844 ground-truth action embeddings, as well as the \textbf{mean Pearson correlation (Pearson)} across embedding dimensions, which reflects how well the learned embedding space preserves semantic similarity.

    \item \textbf{Semantic Text QA}: We measure the quality of generated answers using three complementary NLP metrics. 
\textbf{BERTScore (BERT)}~\cite{bertscore} computes semantic similarity using contextualized token embeddings from a pretrained BERT model, capturing paraphrases and fine-grained meaning. 
\textbf{BLEU}~\cite{Papineni2002BleuAM} evaluates n\!-\!gram precision between predictions and references, reflecting lexical overlap. 
\textbf{ROUGE-L (ROUGE)}~\cite{Lin2004ROUGEAP} measures the longest common subsequence between texts, capturing phrase-level recall. 
Together, these metrics assess both semantic fidelity and surface-form similarity between predicted answers and ground-truth explanations.

\end{itemize}

\noindent\textbf{Results Analysis.}
As shown in Table~\ref{tab:reasoning_scale_qa}, the models achieve strong textual QA performance (BERTScore $\approx$0.92, ROUGE$\approx$0.49) and moderate but meaningful semantic alignment (R@3 up to 11\% and Pearson up to 0.26), providing reliable semantic grounding despite the large action-embedding space consists of 2844 samples.

\noindent Table~\ref{tab:reasoning_scale_qa} also shows that increasing reasoning-module capacity improves both waypoint grounding and semantic understanding.  
For spatial waypoint prediction, Qwen3-VL~4B achieves the best overall accuracy, obtaining the lowest averaged location error (0.223\,m) and the lowest rotation error (40.89°) across all models.  
Qwen3-VL~8B further stabilizes spatial performance, achieving similarly strong location and rotation errors, indicating that spatial grounding saturates around the 4B–8B scale for Qwen3-VL.  
In contrast, semantic embedding alignment exhibits a different scaling trend: Qwen2.5-VL~7B reaches the highest R@3 and Pearson correlation, demonstrating the strongest alignment between reasoning tokens and action semantics.  
Smaller Qwen3-VL models (2B and 4B) lag in semantic alignment despite strong spatial accuracy, suggesting that \emph{fine-grained action–semantic grounding is more capacity-dependent than geometric waypoint prediction}.  
Overall, scaling improves all metrics, but spatial accuracy peaks earlier (at 4B), whereas semantic–action alignment continues improving with larger reasoning capacity.

\begin{table*}[t]
\centering
\setlength{\tabcolsep}{3pt}
\begin{tabular}{cc|cccc|cccc}
\toprule
\multirow{2}{*}{Model} & \multirow{2}{*}{Params} &
\multicolumn{4}{c|}{EgoMAN Unseen} &
\multicolumn{4}{c}{HOT3D OOD} \\
& &
ADE$\downarrow$ & FDE$\downarrow$ & DTW$\downarrow$ & Rot$^\circ\downarrow$ &
ADE$\downarrow$ & FDE$\downarrow$ & DTW$\downarrow$ & Rot$^\circ\downarrow$ \\
\midrule
Qwen2.5-VL & 3B  
& 0.128 & 0.184 & 0.115 & 33.16 
& 0.146 & 0.221 & 0.135 & 35.82 \\

Qwen2.5-VL & 7B  
& {0.124} & {0.179} & \underline{0.111} & {32.75}
& {0.141} & {0.217} & 0.130 & {35.09} \\

\midrule

Qwen3-VL   & 2B  
& 0.130 & 0.186 & 0.118 & 33.48
& 0.142 & 0.216 & 0.132 & 35.62 \\

Qwen3-VL   & 4B  
& \underline{0.123} & \underline{0.178} & \underline{0.111} & \underline{32.63}
& \textbf{0.139} & \textbf{0.212} & \textbf{0.128} & \underline{34.65} \\

Qwen3-VL   & 8B  
& \textbf{0.122} & \textbf{0.177} & \textbf{0.110} & \textbf{32.31}
& \underline{0.140} & \underline{0.214} & \underline{0.129} & \bf 34.62 \\
\bottomrule
\end{tabular}
\caption{\textbf{Effect of Reasoning Module scale on trajectory prediction.}
Best results are \textbf{bolded} and second-best are \underline{underlined}. Larger reasoning models produce consistently more accurate 6-DoF trajectories on both EgoMAN Unseen and HOT3D OOD, with Qwen3-VL scaling smoothly and the 4B model offering an excellent speed–accuracy trade-off.}
\label{tab:reasoning_scale_traj}
\end{table*}

\subsection{Trajectory Prediction on EgoMAN-Bench}
\label{sec:reason_scale_traj}
\textbf{Evaluation Metrics.}
We measure stage-aware 6-DoF trajectory prediction using four metrics that are consistent with the metrics we use in the main paper:
\begin{itemize}[leftmargin=5mm]

    \item \textbf{ADE}: 
    \textbf{Average Displacement Error (ADE)} is the mean Euclidean distance between the predicted and ground-truth 3D wrist positions over all future timesteps. 
    \item \textbf{FDE}: 
    \textbf{Final Displacement Error (FDE)} measures this distance only at the final prediction timestep. 
    Both are computed in meters and evaluate the overall spatial accuracy and final-state consistency of the trajectory.

    \item \textbf{DTW}: 
    \textbf{Dynamic Time Warping (DTW)} measures the minimum alignment cost between the predicted and ground-truth trajectories after allowing temporal stretching or compression. 
    It captures discrepancies in both spatial shape and temporal progression, making it sensitive to timing errors such as early or late motion onset.

    \item \textbf{Rotation Error (Rot)}: 
    The mean geodesic rotation error computed from the relative rotation matrix between predicted and ground-truth wrist orientations. 
    We use the standard geodesic distance in degrees, which measures the smallest 3D rotational difference between two orientations.

\end{itemize}

\noindent All results use best-of-$K$ (sampling $K=10$) on EgoMAN Unseen and HOT3D OOD)

\noindent\textbf{Results.}
Table~\ref{tab:reasoning_scale_traj} shows that scaling the Reasoning Module leads to consistent improvements in 6-DoF trajectory prediction across both EgoMAN Unseen and HOT3D OOD.  
Within each model family, larger variants reduce ADE/FDE and DTW, indicating more accurate and temporally aligned motion forecasts.  
Qwen3-VL~4B and 8B achieve the strongest overall performance, obtaining the lowest ADE, FDE, and rotation errors on EgoMAN Unseen, and competitive or best results on HOT3D OOD.   
Although Qwen2.5-VL~7B maintains solid performance, the Qwen3-VL models benefit more directly from scale, suggesting that \emph{trajectory prediction, unlike semantic embedding alignment, scales smoothly and saturates later in the Qwen3 family}.

\noindent Overall, increasing reasoning-module capacity strengthens stage-aware 6-DoF  trajectory prediction. 
From a speed–performance perspective, Qwen3-VL~4B provides an excellent balance between efficiency and accuracy, while the 7B–8B models offer the strongest overall trajectory quality at higher computational cost.

\section{Prompt Examples}
\label{sec:prompt}
In this section, we detail the LLM-based prompting pipeline used to construct the EgoMAN benchmark. Our pipeline consists of four stages: (i) extracting fine-grained hand--object interaction segments with temporal structure, (ii) filtering invalid or irrelevant interactions and canonicalizing intention goals, (iii) generating diverse QA pairs for reasoning pre-training, and (iv) filtering trajectory phrases to retain only visually grounded, unambiguous interaction samples. The corresponding prompts are shown in Figs.~\ref{fig:prompt1}--\ref{fig:prompt4}.

\subsection{EgoMAN Interaction Annotation}
We first extract temporally localized, atomic hand--object interactions from continuous egocentric video. Given reference narrations and timestamps, the LLM is prompted to decompose an interaction into structured \emph{approach} and \emph{manipulation} stages, each annotated with start/end times, coarse trajectory attributes (start/end locations and shape), and a natural-language atomic description and reasoning. The output is serialized as JSON and forms the core interaction representation used in later stages (Fig.~\ref{fig:prompt1}).

\subsection{Valid Interaction Annotation Filtering}
Not all extracted segments correspond to clean, usable hand--object interactions. We therefore perform a second LLM pass to filter invalid or noisy annotations. Given a candidate atomic description and the reference annotations, the model judges whether the interaction is (i) relevant to the underlying sequence, and (ii) a true hand--object interaction performed by the main subject, rather than background motion or non-manipulative activities. It also summarizes the high-level intention goal into a short phrase, which we later use as a canonical intent label and conditioning token (Fig.~\ref{fig:prompt2}).

\subsection{EgoMAN QA Generation}
For each valid interaction, we generate a set of diverse, intent-aware QA pairs used to train the Reasoning Module. The prompt in Fig.~\ref{fig:prompt3} guides the LLM to produce 8--12 short question--answer pairs that cover complementary aspects of the next interaction: intention goal, which hand will be used, upcoming action and object, spatial trajectory, temporal onset and completion, atomic motion description, and causal reasoning (``why'' the action occurs). The prompt enforces that all answers must be grounded strictly in the provided interaction data and that the intention goal is injected in multiple phrasings to encourage robust semantic alignment.

\subsection{Trajectory Filtering}
We apply an image-conditioned filtering step to ensure that the interaction phrases used for trajectory prediction are visually grounded and unambiguous. As shown in Fig.~\ref{fig:prompt4}, the LLM is asked to verify that (i) the described interaction is physically realistic, (ii) the target object is clearly visible in the egocentric frame, (iii) the image quality is sufficient, and (iv) the phrase refers to a single, unambiguous object. The model outputs a binary validity flag and a short failure reason when rejected. This step prunes low-quality or ambiguous samples and improves the reliability of EgoMAN-Bench trajectory supervision.

\section{Limitations and Future Work}
\label{sec:limitations}
While EgoMAN demonstrates strong intention-conditioned 6-DoF trajectory prediction, several limitations remain. First, our modeling focuses primarily on wrist-level 6-DoF motion and considers only coarse interaction stages (\texttt{<START>}, \texttt{<CONTACT>}, \texttt{<END>}). More fine-grained sub-stages—such as pre-contact adjustment, micro-corrections during manipulation, or multi-step object reorientation—are not explicitly modeled, limiting the system’s ability to capture high-resolution dexterous behavior. Second, although our dataset is large-scale and diverse, it inevitably contains sensor noise, imperfect annotations, and no human verification loop; higher-quality 3D trajectories and cleaner supervision would further benefit learning.

Future work includes extending the representation from wrist trajectories to full hand pose and articulation, enabling more fine-grained reasoning about object manipulation and grasp dynamics. Incorporating multi-stage interaction parsing and richer contact semantics would further improve temporal grounding. Improving dataset quality through higher-fidelity 3D annotations or curated human-verified demonstrations could significantly enhance supervision for fine-grained manipulation learning. Finally, deploying EgoMAN-derived policies on real robotic systems presents an exciting direction for evaluating how intention-grounded 6-DoF predictions transfer to embodied manipulation performance.

\begin{figure*}[htb]
\centering
\begin{tcolorbox}[
    enhanced,
    colback=gray!2!white,
    colframe=black!30,
    title={\textbf{Prompt: Hand–Object Interaction Extraction}},
    fonttitle=\bfseries,
    left=4mm, right=4mm, top=2mm, bottom=2mm,
    boxrule=0.4pt, arc=2pt
]
\small
\noindent
\textbf{System Instruction:} Extract hand–object interactions from video frames.

\vspace{1.5mm}
\textbf{Reference Atomic Description with Timestamps:} \texttt{\{ref\_annos\}}

\vspace{1.5mm}
\textbf{Output Format (JSON):}
\begin{quote}
\ttfamily
\{
\\\hspace*{4mm}"intent": "<action\_goal>",\\
\hspace*{4mm}"interactions": [\\
\hspace*{8mm}\{\\
\hspace*{12mm}"approach": \{\\
\hspace*{16mm}"start\_time": <float>,\\
\hspace*{16mm}"end\_time": <float>,\\
\hspace*{16mm}"trajectory": \{\\
\hspace*{20mm}"start\_point": "<location>",\\
\hspace*{20mm}"end\_point": "<location>",\\
\hspace*{20mm}"shape": "<linear|curved|arc>"\\
\hspace*{16mm}\}\\
\hspace*{12mm}\},\\[1mm]
\hspace*{12mm}"manipulation": \{\\
\hspace*{16mm}"start\_time": <float>,\\
\hspace*{16mm}"end\_time": <float>,\\
\hspace*{16mm}"verb": "<action>",\\
\hspace*{16mm}"object": "<object with short appearance description not mention hand>",\\
\hspace*{16mm}"hand": "<left|right|both>",\\
\hspace*{16mm}"trajectory": \{\\
\hspace*{20mm}"start\_point": "<location>",\\
\hspace*{20mm}"end\_point": "<location>",\\
\hspace*{20mm}"shape": "<linear|curved|arc>"\\
\hspace*{16mm}\}\\
\hspace*{12mm}\},\\[1mm]
\hspace*{12mm}"atomic\_description": "<interaction description>",\\
\hspace*{12mm}"reasoning": "<why the action serves the goal and the trajectory pattern>"\\
\hspace*{8mm}\}\\
\hspace*{4mm}]\\
\}

\end{quote}

\vspace{1.5mm}
\textbf{Rules:}
\begin{itemize}[leftmargin=5mm, itemsep=0.4mm]
    \item The \texttt{approach} stage exists when the hand moves to reach the manipulation location; there is no contact until the object is touched.
    \item If the hand is already in contact at the manipulation location, skip the approach stage and start with \texttt{manipulation}.
    \item Each stage’s trajectory must include three keys: \texttt{start\_point}, \texttt{end\_point}, and \texttt{shape} (one-word movement pattern).
    \item \textbf{Short reasoning:} Explain both (1) why the action serves the overall intent and (2) why the trajectory follows this pattern.
    \item Use precise timestamps derived from video frames.
\end{itemize}

\end{tcolorbox}
\vspace{-3mm}
\caption{Prompt used to generate fine-grained hand–object interaction annotations.}
\label{fig:prompt1}
\end{figure*}

\begin{figure*}[htb]
\centering
\begin{tcolorbox}[
    enhanced,
    colback=gray!2!white,
    colframe=black!30,
    title={\textbf{Prompt: Interaction Validity Judgment}},
    fonttitle=\bfseries,
    left=4mm, right=4mm, top=2mm, bottom=2mm,
    boxrule=0.4pt, arc=2pt
]
\small
\noindent
\textbf{System Instruction:} Judge whether a proposed hand--object interaction is valid and relevant to the reference atomic descriptions, and summarize the high-level intention goal.

\vspace{1.5mm}
\textbf{Inputs:}\\
\texttt{interact\_data.atomic\_description} \quad\; {(free-text for the candidate interaction)}\\
\texttt{\{ref\_annos\}} \quad\; {(reference atomic descriptions with timestamps)}

\vspace{1.5mm}
\textbf{Decision Criteria:}
\begin{itemize}[leftmargin=5mm, itemsep=0.4mm]
    \item \textbf{Relevance:} The reference atomic descriptions summarize the whole action sequence. If the candidate interaction could plausibly be a sub-stage or part of this sequence, consider it relevant unless there is an obvious contradiction.
    \item \textbf{Validity:} The interaction must involve an actual hand--object interaction by the subject C (not just moving in air, not other participants, not whole-body locomotion, not looking-only).
    \item \textbf{Intention Goal:} Provide a concise, high-level phrase describing the goal of the interaction. Do not use parentheses or slashes.
\end{itemize}

\vspace{1.5mm}
\textbf{Output Format (JSON):}
\begin{quote}
\ttfamily
\{\\
\hspace*{4mm}"valid": "<valid|invalid>",\\
\hspace*{4mm}"intention\_goal": "<short\_high\_level\_goal\_phrase>"\\
\}
\end{quote}

\vspace{1.5mm}
\textbf{System Prompt (for LLMs):}
\begin{quote}
\small
\ttfamily
You are an expert in analyzing hand-object interactions. Given an interaction description and reference atomic descriptions, judge: 1) relevance to the reference, 2) whether it is a real hand-object interaction of subject C, and 3) summarize the high-level intention goal.\\
Return ONLY a JSON object with keys "valid" and "intention\_goal".
\end{quote}

\end{tcolorbox}
\vspace{-3mm}
\caption{Prompt used to judge interaction validity and summarize the intention goal.}
\label{fig:prompt2}
\end{figure*}

\begin{figure*}[htb]
\centering
\begin{tcolorbox}[
    enhanced,
    colback=gray!2!white,
    colframe=black!30,
    title={\textbf{Prompt: QA Generation}},
    fonttitle=\bfseries,
    left=4mm, right=4mm, top=2mm, bottom=2mm,
    boxrule=0.4pt, arc=2pt
]
\small
\noindent
\textbf{System Instruction:} Generate short, diverse question--answer pairs about the next hand--object interaction using \textbf{only} the provided data.

\vspace{1.5mm}
\textbf{Input:} \texttt{\{interact\_data\}} {(represents the next interaction to predict)}

\vspace{1.5mm}
\textbf{Rules:}
\begin{itemize}[leftmargin=5mm, itemsep=0.4mm]
    \item Answers must be short natural phrases. \textbf{Only use information from the provided data} --- do not fabricate details such as timing.
    \item \textbf{Do not use parentheses in questions or answers.}
    \item Generate \textbf{8--12} QA pairs covering:
    \begin{enumerate}[leftmargin=6mm, itemsep=0.2mm]
        \item One question asking about the current intention goal.
        \item For other questions, inject the intention goal in diverse formats, e.g.:\\
        \quad ``Given the intention to \ldots'', ``To achieve \ldots, \ldots?'', ``While pursuing \ldots, \ldots?'',\\
        \quad ``In order to \ldots, \ldots?'', ``When attempting to \ldots, \ldots?'', ``For the purpose of \ldots, \ldots?'',\\
        \quad ``As part of \ldots, \ldots?'', ``\ldots to accomplish \ldots?''
        \item Which hand will be used next.
        \item What action will occur next.
        \item What object will be manipulated next.
        \item What trajectory the hands will follow next for manipulation.
        \item When the next manipulation will start/end or start and end ({use manipulation timestamps}).
        \item Where the next manipulation will start/end or start and end.
        \item What is the next atomic motion (the atomic description of the next interaction).
        \item Why the next action will happen (reasoning).
        \item \textbf{If an approach stage exists before manipulation:} when the approach ends; where the approach starts/ends; what the approach trajectory is like.
    \end{enumerate}
\end{itemize}

\vspace{1.5mm}
\textbf{Output Format (JSON Array):}
\begin{quote}
\ttfamily
[\\
\hspace*{4mm}\{ "q": "<question\_text>", "a": "<short\_answer>" \},\\
\hspace*{4mm}\{ "q": "<question\_text>", "a": "<short\_answer>" \},\\
\hspace*{4mm}\ldots\\
]
\end{quote}

\vspace{1.5mm}
\textbf{System Prompt (for LLMs):}
\begin{quote}
\small
\ttfamily
You are a data annotator. Generate diverse QA pairs about the provided hand-object interaction (the next interaction to predict). Follow the Rules exactly and use only the provided data. Output a JSON array of objects with keys "q" and "a".
\end{quote}

\end{tcolorbox}
\vspace{-3mm}
\caption{Prompt used to generate diverse QA pairs for the future hand--object interaction.}
\label{fig:prompt3}
\end{figure*}

\begin{figure*}[htb]
\centering
\begin{tcolorbox}[
    enhanced,
    colback=gray!2!white,
    colframe=black!30,
    title={\textbf{Prompt: Interaction Trajectory Quality Filtering}},
    fonttitle=\bfseries,
    left=4mm, right=4mm, top=2mm, bottom=2mm,
    boxrule=0.4pt, arc=2pt
]
\small
\noindent
\textbf{System Instruction:} Validate whether a proposed interaction phrase is realistic, unambiguous, and visually grounded in the given image.

\vspace{1.5mm}
\textbf{Inputs:}\\
\texttt{image} \quad\; {(single egocentric frame)}\\
\texttt{phrase} \quad\; {(interaction phrase)}

\vspace{1.5mm}
\textbf{Validation Criteria (ALL must be satisfied):}
\begin{itemize}[leftmargin=5mm, itemsep=0.4mm]
    \item \textbf{Realism:} The phrase must describe a physically plausible hand--object interaction.
    \item \textbf{Object Visibility:} The described target object must be clearly visible in the image (hand visibility not required).
    \item \textbf{Image Quality:} The image must be sufficiently clear to identify the target object.
    \item \textbf{Unambiguous Target:} The phrase must specify a single object without ambiguity.
\end{itemize}

\vspace{1.5mm}
\textbf{Output Format (JSON Only):}
\begin{quote}
\ttfamily
\{\\
\hspace*{4mm}"valid": true\\
\}\\[2mm]
or\\[2mm]
\{\\
\hspace*{4mm}"valid": false,\\
\hspace*{4mm}"reason": "<failed\_criterion>"\\
\}
\end{quote}

\vspace{1.5mm}
\textbf{System Prompt (for LLMs):}
\begin{quote}
\small
\ttfamily
Analyze the image together with the interaction phrase: "<PHRASE>".\\
Check whether all four criteria are satisfied: (1) realistic interaction, (2) target object visible, (3) image quality sufficient, (4) target unambiguous.\\
Return ONLY valid JSON:\\
\{\,"valid": true\} if all pass; otherwise \{\,"valid": false, "reason": "<which criterion failed>"\}.\\
Example reasons: "object not visible", "ambiguous target", "poor image quality", "unrealistic interaction".
\end{quote}

\end{tcolorbox}
\vspace{-3mm}
\caption{Prompt used to filter out high-quality interaction trajectory samples.}
\label{fig:prompt4}
\end{figure*}

\end{document}